\pdfoutput=1
\documentclass[runningheads]{llncs}
\usepackage{graphicx}
\usepackage{color}
\usepackage{amsmath,amssymb}
\usepackage{color}

\usepackage{bm}
\usepackage[font=small,labelfont=bf,labelsep=period]{caption}
\usepackage[hidelinks,bookmarks=false,colorlinks=true]{hyperref}
\usepackage{pdfpages}

\makeatletter
\let\MYcaption\@makecaption
\makeatother
\usepackage[labelfont=bf,font=footnotesize,subrefformat=parens]{subcaption}
\makeatletter
\let\@makecaption\MYcaption
\makeatother

\newcommand{\etal}{et al.}
\newcommand{\ie}{i.e., }
\newcommand{\eg}{e.g., }

\DeclareMathOperator*{\argmin}{arg\,min}

\begin{document}

\pagestyle{headings}
\mainmatter

\def\ACCV18SubNumber{672}

\title{Scale Estimation of Monocular SfM\\for a Multi-modal Stereo Camera}
\titlerunning{Scale Estimation of Monocular SfM for a Multi-modal Stereo Camera}

\author{
Shinya Sumikura\inst{1}
\and
Ken Sakurada\inst{2}
\and \\
Nobuo Kawaguchi\inst{1}
\and
Ryosuke Nakamura\inst{2}
}
\authorrunning{S. Sumikura et al.}

\institute{
Nagoya University, Japan\\
\email{sumikura@ucl.nuee.nagoya-u.ac.jp, kawaguti@nagoya-u.jp}
\and
National Institute of Advanced Industrial Science and Technology, Japan\\
\email{\{k.sakurada,r.nakamura\}@aist.go.jp}
}

\maketitle

\begin{abstract}
This paper proposes a novel method of estimating the absolute scale of monocular SfM for a multi-modal stereo camera.
In the fields of computer vision and robotics, scale estimation for monocular SfM has been widely investigated in order to simplify systems.
This paper addresses the scale estimation problem for a stereo camera system in which two cameras capture different spectral images (\eg RGB and FIR), whose feature points are difficult to directly match using descriptors.
Furthermore, the number of matching points between FIR images can be comparatively small, owing to the low resolution and lack of thermal scene texture.
To cope with these difficulties, the proposed method estimates the scale parameter using batch optimization, based on the epipolar constraint of a small number of feature correspondences between the invisible light images.
The accuracy and numerical stability of the proposed method are verified by synthetic and real image experiments.
\end{abstract}

\section{Introduction}
\label{sec:introduction}

\begin{figure}[t]
    \centering
    \includegraphics[width=12.0cm]{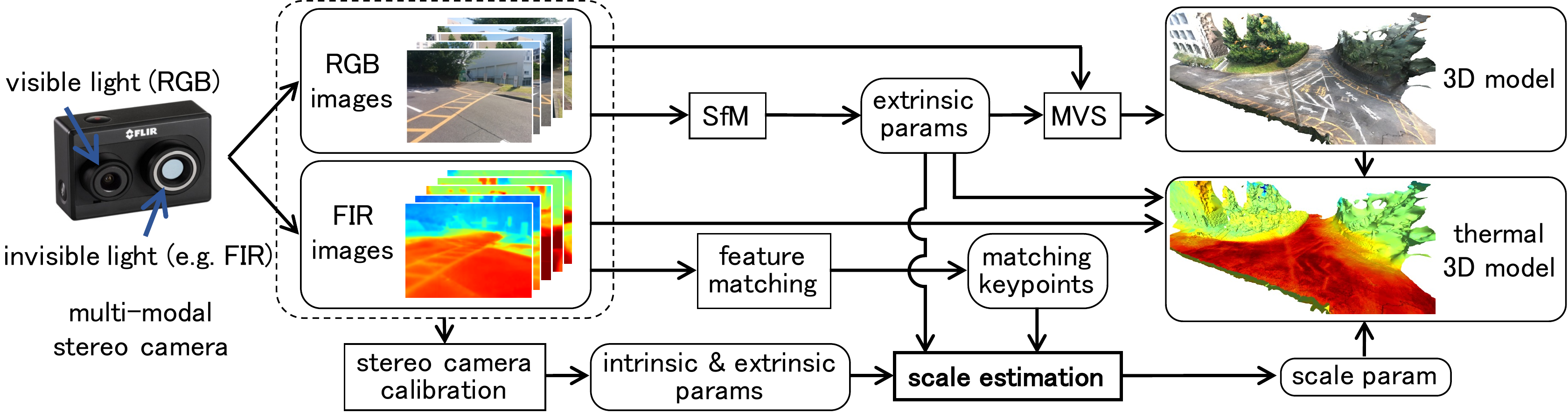}
    \caption{Flowchart of the proposed scale estimation and the application example: thermal 3D reconstruction.}
    \label{fig:flowchart}
\end{figure}

This paper addresses the problem of estimating the scale parameter of monocular Structure from Motion (SfM) for a multi-modal stereo camera system (Fig.\,\ref{fig:flowchart}).
There has been growing interest in scene modeling with the development of mobile digital devices.
In particular, researchers in the field of computer vision and robotics have exhaustively investigated scale estimation methods for monocular SfM to benefit from the simplicity of the camera system~\cite{MonoSLAM,klein2007parallel}.
There are several ways to estimate the scale parameter --- for example, integration with other sensors such as inertial measurement units (IMUs)~\cite{Nuetzi2011} or navigation satellite systems (NSSs), such as the Global Positioning System (GPS).
Also, some methods utilize the prior knowledge of the sensor setups~\cite{Kitt20117357,scaramuzza2009absolute}.
In this paper, the scale parameter of monocular SfM is estimated by integrating the information of different spectral images, such as those taken by RGB and far-infrared (FIR) cameras in a stereo camera setup, whose feature points are difficult to directly match by using descriptors (\eg SIFT~\cite{lowe2004distinctive}, SURF~\cite{bay2006surf}, and ORB~\cite{Rublee2011}).

\begin{figure}[t]
    \centering
    \begin{minipage}[b]{0.485\hsize}
        \centering
        \includegraphics[width=5.5cm]{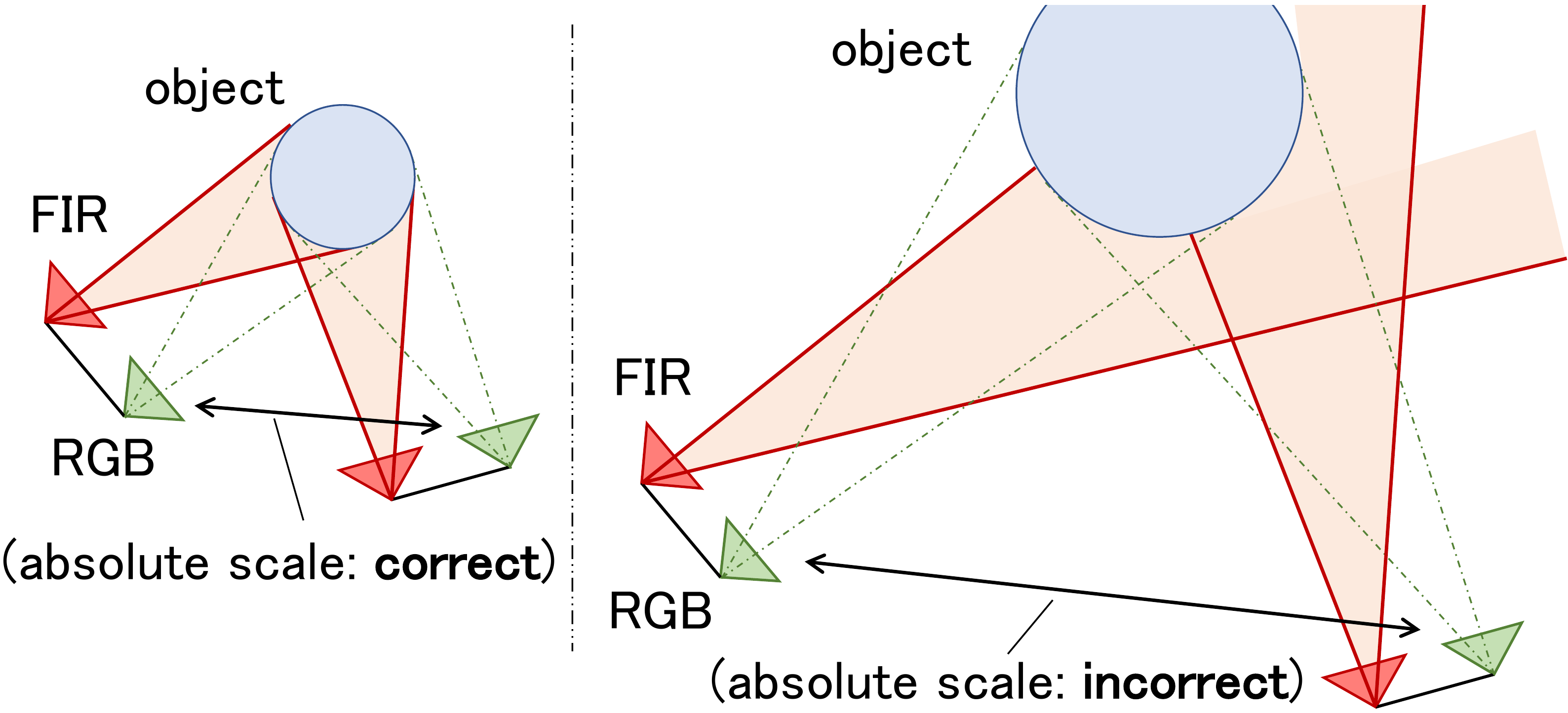}
        \subcaption{Diagrams of thermal projection}
        \label{fig:two_view_geometry_example}
    \end{minipage}
    \begin{minipage}[b]{0.485\hsize}
        \centering
        \includegraphics[width=5.5cm]{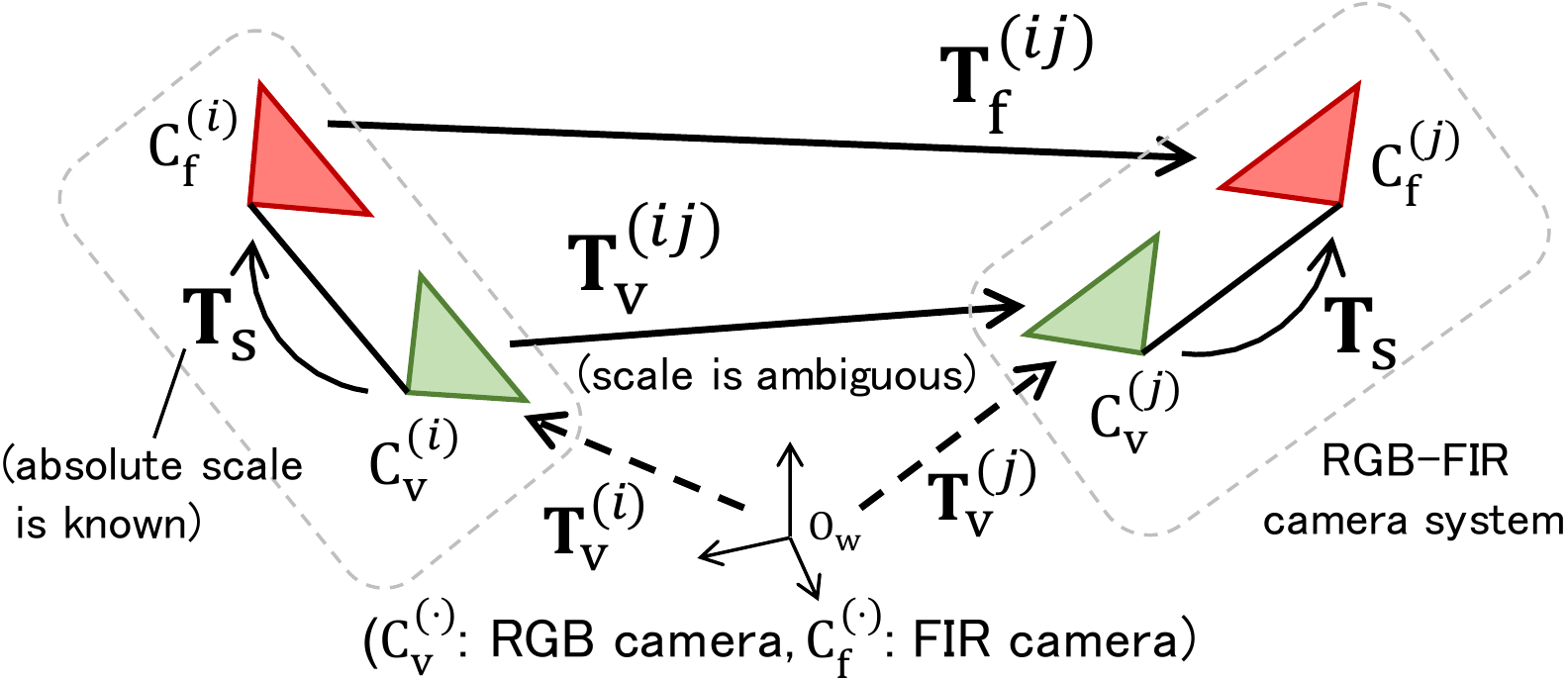}
        \subcaption{Definition of camera poses}
        \label{fig:two_view_geometry}
    \end{minipage}
    \caption{\textbf{(a)} Examples of projection when the absolute scale of RGB camera poses is correct (\textit{left}) or incorrect (\textit{right}).
    Green and red lines indicate the projection of the object in the RGB and FIR images, respectively.
    When the scale is incorrect, the reprojection of the FIR images is misaligned with the object.
    \textbf{(b)} Definition of camera poses for the $i^{\mathrm{th}}$ and $j^{\mathrm{th}}$ viewpoints.
    ${\bf T}_{\rm v}^{(\cdot)}$, ${\bf T}_{\rm f}^{(\cdot)}$ and ${\bf T}_{\rm s}$ represent the global poses of the RGB camera ${\rm C}_{\rm v}^{(\cdot)}$, FIR camera ${\rm C}_{\rm f}^{(\cdot)}$, and the relative pose between them, respectively. ${\bf T}_{(\cdot)}^{(ij)}$ represents the relative pose between the same type of cameras, ${\rm C}_{(\cdot)}^{(i)}$ and ${\rm C}_{(\cdot)}^{(j)}$.}
\end{figure}

With the development of the production techniques of FIR cameras, they have been widely utilized for deriving the benefits of thermal information in the form of infrared radiation emitted by objects, such as infrastructure inspection~\cite{HAM2013395,iwaszczuk2017camera,7676356,vidas20133d,weinmann2014thermal}, pedestrian detection in the dark~\cite{BERTOZZI2007194}, and monitoring volcanic activity~\cite{THIELE2017140}.
Especially for unmanned aerial vehicles (UAVs), a stereo pair of RGB and FIR cameras, which we call a \emph{multi-modal stereo camera}, is often mounted on the UAV for such inspection and monitoring.
Although the multi-modal stereo camera can capture different spectral images simultaneously, for example, in the case of structural inspection, it is labor-intensive to compare a large number of image pairs.
To improve the efficiency of the inspection, SfM~\cite{Agarwal2009,schoenberger2016sfm} and Multi-View Stereo (MVS)~\cite{Furukawa2010,jancosek2011multi,schoenberger2016mvs} can be used for \emph{thermal 3D reconstruction} (Fig.\,\ref{fig:flowchart}).
The estimation of the absolute scale of the monocular SfM is needed in order to project FIR image information to the 3D model (Fig.\,\ref{fig:two_view_geometry_example}).
However, it is difficult to match feature points between RGB and FIR images directly.
Moreover, the number of matching points between FIR images is comparatively small due to the low resolution and the lack of thermal texture in a scene.
Although machine learning methods, such as deep neural networks (DNNs)~\cite{detone2017toward,han2015matchnet,Zagoruyko_2015_CVPR}, can be used to match feature points between different types of images, the cost of dataset creation for every camera and scene is quite expensive.

To estimate the scale parameter from only the information of the multi-modal camera system, we leverage the stereo setup with a constant extrinsic parameter and a small number of feature correspondences between the same modal images other than the visible ones (Fig.\,\ref{fig:flowchart}).
More concretely, the proposed method is based on a least-squares method of residuals by the epipolar constraint between the same modal images.
The main contribution of this paper is threefold:
first, the formulation of the scale estimation for a multi-modal stereo camera system;
second, the verification of the effectiveness of the formulation through synthetic and real image experiments;
and third, experimental thermal 3D mappings as one of the applications of the proposed method.

\section{Related work}
\label{sec:related_work}

\subsection{Thermal 3D reconstruction}

The FIR camera is utilized with other types of sensors for thermal 3D reconstruction because the texture of FIR images is poorer than that of visible ones, especially for indoor scenes.
Oreifej~\etal~\cite{cramer2014automatic} developed a fully automatic 3D thermal mapping system for building interiors using light detection and ranging (LiDAR) sensors to directly measure the depth of a scene.
Additionally, depth image sensors are utilized to estimate the dense 3D model of a scene based on the Kinect Fusion algorithm~\cite{newcombe2011kinectfusion} in the works of~\cite{7676356,vidas20133d}.

A combination of SfM and MVS is an alternative method for the 3D scene reconstruction.
Ham~\etal~\cite{HAM2013395} developed a method to directly match feature points between RGB and FIR images, which works only in rich thermal-texture environments.
Under similar conditions, the method proposed by Truong~\etal~\cite{Truong_2017_ICCV} performs SfM using each of RGB and FIR images independently, aligning the two sparse point clouds.

Whereas the measurement range of the LiDAR sensor is longer than that of the depth image sensor, it has disadvantages in sensor size and weight, and is more expensive compared to RGB and depth cameras.
Additionally, the depth image sensor can directly obtain dense 3D point clouds of a scene; however, it is unsuitable for wide-area measurement tasks because the measurement range is comparatively short.
As mentioned, this study assumes thermal 3D reconstruction of wide areas for structural inspection by UAVs as an application.
Thus, this paper proposes a scale estimation method of monocular SfM for a multi-modal stereo camera with the aim of thermal 3D reconstruction using an RGB--FIR camera system.

\subsection{Scale estimation for monocular SfM}
\label{subsec:scale_estimation_for_monocular_sfm}

There are several types of scale estimation methods for monocular SfM based on other sensors and prior knowledge.

To estimate the absolute scale parameter of monocular SfM, an IMU is utilized as an internal sensor to integrate the information of the accelerations and angular velocities with vision-based estimation using the extended Kalman filter (EKF)~\cite{Nuetzi2011}.
As an external sensor, location information from NSSs (\eg GPS) can be used to estimate the similarity transformation between the trajectories of monocular SfM and the GPS information based on a least-squares method.

Otherwise, prior knowledge of the sensor setups is utilized for scale estimation.
Scaramuzza~\etal~\cite{scaramuzza2009absolute} exploit the nonholonomic constraints of a vehicle on which a camera is mounted.
The work by Kitt~\etal~\cite{Kitt20117357} utilizes ground planar detection and the height from the ground of a camera.

The objective of this study is to estimate the scale parameter of monocular SfM from only multi-modal stereo camera images without other sensor information, for versatility.
For example, in the case of structural inspection using UAVs, IMUs mounted on the drones suffer from vibration noise, and the GPS signal cannot be received owing to the structure.
Additionally, assumptions of sensor setups restrain the application of scale estimation.
Therefore, the proposed method utilizes only input image information and pre-calibration parameters.

As one of the scale estimation methods for a multi-modal stereo camera, which uses the information only from such a camera system, Truong~\etal~\cite{Truong_2017_ICCV} proposed a method based on an alignment of RGB and FIR point clouds.
This method requires the point cloud created only from FIR images.
Thus, it is not applicable to scenes with non-rich thermal texture, such as indoor scenes.
Otherwise, considering a multi-modal stereo camera as a multi-camera cluster with non-overlapping fields of view, we can theoretically apply scale estimation methods of monocular SfM for such a multi-camera cluster to a multi-modal stereo camera.
The work by Clipp~\etal~\cite{clipp2008robust} estimates the absolute scale of monocular SfM for a multi-camera cluster with non-overlapping fields of view by minimizing the residual based on the epipolar constraint between two viewpoints.
This method does not perform the batch optimization, which utilizes multiple image pairs, and does not take the scale parameter into account when performing the bundle adjustment (BA)~\cite{triggs1999bundle}.

Thus, in this paper, we compare the proposed scale estimation method with the ones of Truong~\etal~\cite{Truong_2017_ICCV} and by Clipp~\etal~\cite{clipp2008robust}.

\section{Scale estimation}
\label{sec:scale_estimation}

\subsection{Problem formulation}

In this section, we describe a novel method of estimating a scale parameter of reconstruction results from monocular SfM.
Here we use a stereo system of RGB and FIR cameras (\ie RGB--FIR) as an example of a multi-modal stereo camera system.
Fig.\,\ref{fig:two_view_geometry} expresses the global and relative transformation matrices of a system composed of two viewpoints with an RGB--FIR camera system.

We start with a given set of RGB images $\bigl\{ {\rm I}_{\rm v}^{(1)}, {\rm I}_{\rm v}^{(2)}, \cdots, {\rm I}_{\rm v}^{(n)} \bigr\}$, and FIR images $\bigl\{ {\rm I}_{\rm f}^{(1)}, {\rm I}_{\rm f}^{(2)}, \cdots, {\rm I}_{\rm f}^{(n)} \bigr\}$, whose $k^{\rm{th}}$ images, ${\rm I}_{\rm v}^{(k)}$ and ${\rm I}_{\rm f}^{(k)}$, are taken simultaneously using an RGB--FIR camera system whose constant extrinsic parameter is
\begin{align}
    {\bf T}_{\rm s}
    =
    \begin{bmatrix}
        {\bf R}_{\rm s} & \; {\bf t}_{\rm s} \\
        {\bm 0}^{\mathsf T} & \; 1
    \end{bmatrix}
    .
    \label{eq:rgb_fir_extrinsic_without_s}
\end{align}
${\bf R}_{\rm s}$ and ${\bf t}_{\rm s}$ represent the rotation matrix and the translation vector between the two cameras of the camera system, respectively.
Those matrix and vector are estimated via calibration in advance.
Additionally, we assume that the $k^{\rm{th}}$ images, ${\rm I}_{\rm v}^{(k)}$ and ${\rm I}_{\rm f}^{(k)}$, are taken by the $k^{\rm{th}}$ cameras, ${\rm C}_{\rm v}^{(k)}$(RGB) and ${\rm C}_{\rm f}^{(k)}$(FIR), with the global extrinsic parameters, ${\bf T}_{\rm v}^{(k)}$ and ${\bf T}_{\rm f}^{(k)}$, respectively.
Note that ${\rm C}_{\rm v}^{(k)}$ and ${\rm C}_{\rm f}^{(k)}$ comprise the pair of cameras in the RGB--FIR camera system.
$\bigl\{ {\bf T}_{\rm v}^{(k)} \bigr\}$ can be estimated except for its absolute scale by monocular SfM of the RGB images.

Using ${\bf T}_{\rm v}^{(i)}$ and ${\bf T}_{\rm v}^{(j)}$, the relative transformation between ${\rm C}_{\rm v}^{(i)}$ and ${\rm C}_{\rm v}^{(j)}$ is computed by ${\bf T}_{\rm v}^{(j)} {{\bf T}_{\rm v}^{(i)}}^{-1}$.
To solve the scale ambiguity, a scale parameter $s \in \mathbb{R}$ is introduced.
Then, the relative transformation ${\bf T}_{\rm v}^{(ij)}$ between ${\rm C}_{\rm v}^{(i)}$ and ${\rm C}_{\rm v}^{(j)}$ including the scale parameter $s$ is expressed by
\begin{align}
    {\bf T}_{\rm v}^{(ij)}
    =
    \begin{bmatrix}
        {\bf R}_{\rm v}^{(ij)} & \; s \cdot {\bf t}_{\rm v}^{(ij)} \\
        {\bm 0}^{\mathsf T} & \; 1
    \end{bmatrix}
    ,
    \label{eq:rgb_rgb_extrinsic_with_s}
\end{align}
where ${\bf R}_{\rm v}^{(ij)}$ and ${\bf t}_{\rm v}^{(ij)}$ are the rotation matrix block and the translation vector block of ${\bf T}_{\rm v}^{(j)} {{\bf T}_{\rm v}^{(i)}}^{-1}$, respectively.
The goal is to estimate the correct $s \in \mathbb{R}$.

\subsection{Derivation of scale parameter $s$}

With ${\bf T}_{\rm v}^{(ij)}$ and ${\bf T}_{\rm s}$, the relative transformation ${\bf T}_{\rm f}^{(ij)} = {\bf T}_{\rm s} {\bf T}_{\rm v}^{(ij)} {\bf T}_{\rm s}^{-1}$ between the two FIR cameras, ${\rm C}_{\rm f}^{(i)}$ and ${\rm C}_{\rm f}^{(j)}$, can be computed as
\begin{align}
    {\bf T}_{\rm f}^{(ij)}
    &=
    \begin{bmatrix}
        {\bf R}_{\rm s} {\bf R}_{\rm v}^{(ij)} {\bf R}_{\rm s}^{-1}
        & \;
        s \cdot {\bf R}_{\rm s} {\bf t}_{\rm v}^{(ij)} + ({\bf I} - {\bf R}_{\rm s} {\bf R}_{\rm v}^{(ij)} {\bf R}_{\rm s}^{-1}) {\bf t}_{\rm s}
        \\
        {\bm 0}^{\mathsf T}
        & \;
        1
    \end{bmatrix}
    \\
    &=
    \begin{bmatrix}
        {\bf A}^{(ij)}
        & \;
        s \cdot {\bf b}^{(ij)} + {\bf c}^{(ij)}
        \\
        {\bm 0}^{\mathsf T}
        & \;
        1
    \end{bmatrix}
    ,
    \label{eq:a_b_c_with_scale_paramerer}
\end{align}
where ${\bf A}^{(ij)} = \bigl[ {\bf a}_1^{(ij)} \big{|}\, {\bf a}_2^{(ij)} \big{|}\, {\bf a}_3^{(ij)} \bigr] = {\bf R}_{\rm s} {\bf R}_{\rm v}^{(ij)} {\bf R}_{\rm s}^{-1}$, ${\bf b}^{(ij)} = {\bf R}_{\rm s} {\bf t}_{\rm v}^{(ij)}$ and ${\bf c}^{(ij)} = ({\bf I} - {\bf R}_{\rm s} {\bf R}_{\rm v}^{(ij)} {\bf R}_{\rm s}^{-1}) {\bf t}_{\rm s}$.
An essential matrix ${\bf E}^{(ij)}$ between ${\rm C}_{\rm f}^{(i)}$ and ${\rm C}_{\rm f}^{(j)}$ can be derived from ${\bf T}_{\rm f}^{(ij)}$ and expressed as
\begin{align}
    {\bf E}^{(ij)}
    &=
    \bigl[ s {\bf b}^{(ij)} + {\bf c}^{(ij)} \bigr]_{\times} {\bf A}^{(ij)}
    \\
    &=
    s \cdot \bigl[ {\bf b}^{(ij)} \times {\bf a}_1^{(ij)} \big{|}\; {\bf b}^{(ij)} \times {\bf a}_2^{(ij)} \big{|}\; {\bf b}^{(ij)} \times {\bf a}_3^{(ij)} \bigr]
    \nonumber
    \\
    &\quad \; \hspace{0.2pt}
    +\bigl[ {\bf c}^{(ij)} \hspace{1.25pt} \times {\bf a}_1^{(ij)} \big{|}\; {\bf c}^{(ij)} \hspace{1.25pt} \times {\bf a}_2^{(ij)} \big{|}\; {\bf c}^{(ij)} \hspace{1.3pt}\times {\bf a}_3^{(ij)} \bigr]
    .
\end{align}
The epipolar constraint between the two FIR images, ${\rm I}_{\rm f}^{(i)}$ and ${\rm I}_{\rm f}^{(j)}$, corresponding to the FIR cameras, ${\rm C}_{\rm f}^{(i)}$ and ${\rm C}_{\rm f}^{(j)}$, is formulated as
\begin{align}
    {{\bf p}_{k}^{(j)}}^{\mathsf T}
    {\bf E}^{(ij)}
    {\bf p}_{k}^{(i)}
    =
    0
    ,
    \label{eq:epipolar_constraint_ij}
\end{align}
where ${\bf p}_{k}^{(i)} = \bigl[ x_k^{(i)},\: y_k^{(i)},\: 1 \bigr]^{\mathsf T}$ and ${\bf p}_{k}^{(j)} = \bigl[ x_k^{(j)},\: y_k^{(j)},\: 1 \bigr]^{\mathsf T}$ are the $k^{\rm{th}}$ corresponding feature points between ${\rm I}_{\rm f}^{(i)}$ and ${\rm I}_{\rm f}^{(j)}$, in the form of normalized image coordinates~\cite{Hartley2004}.
A normalized image point ${\bf p}_{\mathit k}^{\mathit (i)}$ is defined as
\begin{align}
    {\bf p}_{\mathit k}^{\mathit (i)}
    =
    {\bf K}_{\rm f}^{-1}
    \begin{bmatrix}
        u_{\mathit k}^{\mathit (i)}
        ,\:
        v_{\mathit k}^{\mathit (i)}
        ,\:
        1
    \end{bmatrix}^{\mathsf T}
    \;\; \textrm{with} \;\;
    {\bf K_{\rm f}}
    =
    \begin{bmatrix}
        f_{x} & \; 0 & \; c_{x} \\
        0 & \; f_{y} & \; c_{y} \\
        0 & \; 0 & \; 1
    \end{bmatrix}
    ,
    \label{eq:convert_image_to_normalized_3D_point}
\end{align}
where ${\bf K_{\rm f}}$ is the intrinsic parameter matrix of the FIR camera.
$\bigl[ u_{\mathit k}^{\mathit (i)}, v_{\mathit k}^{\mathit (i)} \bigr]$ is the feature point in pixels in ${\rm I}_{\rm f}^{(i)}$ and is the $k^{\rm{th}}$ corresponding feature point with $\bigl[ u_{\mathit k}^{\mathit (j)}, v_{\mathit k}^{\mathit (j)} \bigr]$ in ${\rm I}_{\rm f}^{(j)}$.
Additionally, the normalized image point is also defined as
\begin{align}
    {\bf p}_{\mathit k}^{\mathit (i)}
    =
    {\bf X}_{\mathit l}^{\mathit (i)} \Big/ Z_l^{(i)}
    ,
    \label{eq:convert_world_to_normalized_3D_point}
\end{align}
where ${\bf X}_{\mathit l}^{\mathit (i)} = [X_l^{(i)}, Y_l^{(i)}, Z_l^{(i)}]^{\mathsf T}$ is the $l^{\rm{th}}$ 3D point in the coordinate system of the $i^{\rm{th}}$ FIR camera ${\rm C}_{\rm f}^{(i)}$.
Here, ${\bf X}_{\mathit l}^{\mathit (i)}$ corresponds to the feature point ${\bf p}_{\mathit k}^{\mathit (i)}$ on ${\rm I}_{\rm f}^{(i)}$.

The epipolar constraint of Equation (\ref{eq:epipolar_constraint_ij}) can be expanded to
\begin{align}
    {\bf u}_{k}^{(ij)} \bigl( s \cdot {\bf f}^{(ij)} + {\bf g}^{(ij)} \bigr) = 0
    \label{eq:epipolar_constraint_expanded}
\end{align}
with
\begin{align}
    {\bf u}_{k}^{(ij)}
    =&\,
    \bigl[
    x_k^{(i)} x_k^{(j)},\; x_k^{(i)} y_k^{(j)},\; x_k^{(i)},\; y_k^{(i)} x_k^{(j)},\; y_k^{(i)} y_k^{(j)},\; y_k^{(i)},\; x_k^{(j)},\; y_k^{(j)},\; 1
    \bigr]
    ,
    \label{eq:u_k_ij}
    \\
    {\bf f}^{(ij)}
    =&\,
    \Bigl[
    \bigl[ {\bf b}^{(ij)} \mathalpha{\times} {\bf a}_1^{(ij)} \bigr]_1,
    \bigl[ {\bf b}^{(ij)} \mathalpha{\times} {\bf a}_1^{(ij)} \bigr]_2,
    \cdots,
    \bigl[ {\bf b}^{(ij)} \mathalpha{\times} {\bf a}_3^{(ij)} \bigr]_2,
    \bigl[ {\bf b}^{(ij)} \mathalpha{\times} {\bf a}_3^{(ij)} \bigr]_3
    \Bigr]^{\mathsf T}
    ,
    \\
    {\bf g}^{(ij)}
    =&\,
    \Bigl[
    \bigl[ {\bf c}^{(ij)} \hspace{0.8pt}\mathalpha{\times}\hspace{0.48pt} {\bf a}_1^{(ij)} \bigr]_1,
    \bigl[ {\bf c}^{(ij)} \hspace{0.8pt}\mathalpha{\times}\hspace{0.48pt} {\bf a}_1^{(ij)} \bigr]_2,
    \cdots,
    \bigl[ {\bf c}^{(ij)} \hspace{0.8pt}\mathalpha{\times}\hspace{0.48pt} {\bf a}_3^{(ij)} \bigr]_2,
    \bigl[ {\bf c}^{(ij)} \hspace{0.8pt}\mathalpha{\times}\hspace{0.48pt} {\bf a}_3^{(ij)} \bigr]_3
    \Bigr]^{\mathsf T}
    .
\end{align}
If the coordinates of the feature points have no error, Equation (\ref{eq:epipolar_constraint_expanded}) is completely satisfied.
However, in reality, the equation is not completely satisfied because coordinates of feature points usually have some error and the scale $s$ is unknown.
In such a case, the scalar residual $e_{k}^{(ij)}$ is defined as
\begin{align}
    e_{k}^{(ij)} = {\bf u}_{k}^{(ij)} \bigl( s \cdot {\bf f}^{(ij)} + {\bf g}^{(ij)} \bigr)
    .
    \label{eq:epipolar_scalar_residual}
\end{align}
Likewise, the residual vector ${\bf e}^{(ij)}$ can be defined by
\begin{align}
    {\bf e}^{(ij)}
    = \;
    &{\bf U}^{(ij)} \bigl( s \cdot {\bf f}^{(ij)} + {\bf g}^{(ij)} \bigr)
    \;\; \mathrm{with} \;\;
    {\bf U}^{(ij)}
    =
    \Bigl[ \;
        {{\bf u}_{1}^{(ij)}}^{\mathsf T} \big{|}\;
        {{\bf u}_{2}^{(ij)}}^{\mathsf T} \big{|}
        \cdots \big{|}\;
        {{\bf u}_{n}^{(ij)}}^{\mathsf T}
    \; \Bigr]^{\mathsf T}
    ,
    \label{eq:epipolar_vector_residual}
\end{align}
where $n$ is the number of corresponding feature points between ${\rm I}_{\rm f}^{(i)}$ and ${\rm I}_{\rm f}^{(j)}$.
Using a least-squares method, the scale parameter $s$ can be estimated by
\begin{align}
   s
   &= \argmin_{s \: \in \: \mathbb{R}} \frac{1}{2} \sum_{i, j, i \neq j} \left|\left| {\bf e}^{(ij)} \right|\right|^2
   .
\end{align}
Collectively, the scale estimation problem comes down to determining $s$, such that the error function,
\begin{align}
    J(s)
    &=
    \frac{1}{2} \sum_{i, j, i \neq j} \left|\left| {\bf U}^{(ij)} \bigl( s \cdot {\bf f}^{(ij)} + {\bf g}^{(ij)} \bigr) \right|\right|^2
\end{align}
is minimized.
Thus, the scale parameter $s$ is determined by solving the equation $\mathrm{d} J(s) / \mathrm{d} s = 0$ in terms of $s$.
Therefore, the scale $s$ is computed by
\begin{align}
    s
    =
    -
    \sum_{i, j, i \neq j} \Bigl( {{\bf f}^{(ij)}}^{\mathsf T} {{\bf U}^{(ij)}}^{\mathsf T} {{\bf U}^{(ij)}} {{\bf g}^{(ij)}} \Bigr)
    \: \bigg/ \:
    \sum_{i, j, i \neq j} \Bigl( {{\bf f}^{(ij)}}^{\mathsf T} {{\bf U}^{(ij)}}^{\mathsf T} {{\bf U}^{(ij)}} {{\bf f}^{(ij)}} \Bigr)
    .
    \label{eq:scale_estimation_equation}
\end{align}

\subsection{Alternative derivation}

In Equation (\ref{eq:rgb_rgb_extrinsic_with_s}), the scale parameter $s$ and the relative translation vector ${\bf t}_{\rm v}^{(ij)}$ between the two RGB cameras, ${\rm C}_{\rm f}^{(i)}$ and ${\rm C}_{\rm f}^{(j)}$, are multiplied.
The scale parameter $s$ can be alternatively applied to the translation vector ${\bf t}_{\rm s}$ in ${\bf T}_{\rm s}$, in contrast to Equations (\ref{eq:rgb_fir_extrinsic_without_s}) and (\ref{eq:rgb_rgb_extrinsic_with_s}).
This introduction of $s$ is reasonable because multiplying ${\bf t}_{\rm s}$ by $s$ is geometrically equivalent to multiplying ${\bf t}_{\rm v}^{(ij)}$ by $1 / s$.
Therefore, we can also estimate the scale parameter of monocular SfM, which has scale ambiguity, from
\begin{align}
    {\bf T}_{\rm v}^{(ij)}
    =
    \begin{bmatrix}
        {\bf R}_{\rm v}^{(ij)} & \; {\bf t}_{\rm v}^{(ij)} \\
        {\bm 0}^{\mathsf T} & \; 1
    \end{bmatrix}
    \quad \mathrm{and} \quad
    {\bf T}_{\rm s}
    =
    \begin{bmatrix}
        {\bf R}_{\rm s} & \; s \cdot {\bf t}_{\rm s} \\
        {\bm 0}^{\mathsf T} & \; 1
    \end{bmatrix}
    \label{eq:alternative_extrinsics}
    .
\end{align}
When using Equation (\ref{eq:alternative_extrinsics}) for scale estimation, the ${\bf A}^{(ij)}$, ${\bf b}^{(ij)}$ and ${\bf c}^{(ij)}$ in Equation (\ref{eq:a_b_c_with_scale_paramerer}) are
\begin{align}
    {\bf A}^{(ij)}
    =
    {\bf R}_{\rm s} {\bf R}_{\rm v}^{(ij)} {\bf R}_{\rm s}^{-1}
    \; \textrm{,} \;\;
    {\bf b}^{(ij)}
    =
    ({\bf I} - {\bf R}_{\rm s} {\bf R}_{\rm v}^{(ij)} {\bf R}_{\rm s}^{-1}) {\bf t}_{\rm s}
    \;\; \textrm{and} \;\;
    {\bf c}^{(ij)}
    =
    {\bf R}_{\rm s} {\bf t}_{\rm v}^{(ij)}
    .
\end{align}
The rest of the derivation procedure remains the same.

Hereinafter, the formula for the scale estimation based on Equations (\ref{eq:rgb_fir_extrinsic_without_s}) and (\ref{eq:rgb_rgb_extrinsic_with_s}) is called Algorithm (1), whereas the formula based on Equation (\ref{eq:alternative_extrinsics}) is called Algorithm (2).

\subsection{Scale-oriented bundle adjustment}

After an initial estimation of the scale parameter by Equation (\ref{eq:scale_estimation_equation}) of Algorithm (1) or (2), we perform the bundle adjustment (BA)~\cite{triggs1999bundle}.
Before the scale estimation, the camera poses of the RGB cameras are precisely estimated via monocular SfM, except for its absolute scale.
Thus, our BA optimizes the scale parameter $s$ rather than the translation vectors of the RGB cameras.

Using the scale parameter $s$, the reprojection error ${\bm \delta}^{(i)}_{k,l}$ of the $l^{\mathrm{th}}$ FIR 3D point ${\bf X}_{l} = [X_{l},\: Y_{l},\: Z_{l}]^{\mathsf T}$ (in the world coordinate system) in the FIR image ${\rm I}_{\rm f}^{(i)}$ is defined as
\begin{align}
    {\bm \delta}^{(i)}_{k,l}
    =
    {\bf x}_{\mathit k}^{\mathit (i)} - \pi^{(i)} \left( s, {\bf X}_{l} \right)
    ,
\end{align}
where ${\bf x}_{\mathit k}^{\mathit (i)}$ represents the $k^{\mathrm{th}}$ feature point in the $i^{\mathrm{th}}$ FIR image ${\rm I}_{\rm f}^{(i)}$ and corresponds to ${\bf X}_{l}$.
The projection function $\pi^{(i)}(\cdot)$ for the $i^{\mathrm{th}}$ FIR camera is
\begin{align}
    \pi^{(i)} \left( s, {\bf X}_{l} \right) = \left[ f_{x} X_{l}^{(i)} \big/ Z_{l}^{(i)} + c_{x}, \; f_{y} Y_{l}^{(i)} \big/ Z_{l}^{(i)} + c_{y} \right]^{\mathsf T}
    ,
\end{align}
where ${\bf X}_{\mathit l}^{\mathit (i)} = [X_l^{(i)},\: Y_l^{(i)},\: Z_l^{(i)}]^{\mathsf T}$ is computed by
\begin{align}
    {\bf X}_{l}^{(i)} &=
    {\bf R}_{\rm s} {\bf R}_{\rm v}^{(i)} {\bf X}_{l} + s \cdot {\bf R}_{\rm s} {\bf t}_{\rm v}^{(i)} + {\bf t}_{\rm s}
    \quad
    \bigl( \textrm{when using Algorithm (1)} \bigr)
    ,
    \\
    {\bf X}_{l}^{(i)} &=
    {\bf R}_{\rm s} {\bf R}_{\rm v}^{(i)} {\bf X}_{l} + {\bf R}_{\rm s} {\bf t}_{\rm v}^{(i)} + s \cdot {\bf t}_{\rm s}
    \quad
    \bigl( \textrm{when using Algorithm (2)} \bigr)
    .
\end{align}
The cost function $\mathcal{L} \left( \cdot \right)$ composed of the reprojection errors is defined by
\begin{align}
    \mathcal{L} \left( s, \big\{ {\bf X}_{l} \big\}, {\bf K_{\rm f}}; \big\{ {\bf T}_{\rm v}^{(i)} \big\}, {\bf T}_{\rm s} \right)
    =
    \sum_{i, k, l} \rho_\mathrm{h} \left( \left|\left| {\bm \delta}^{(i)}_{k,l} \right|\right|^2 \; \Big/ \sigma_{\mathrm{r}}^2 \right)
    ,
\end{align}
where $\rho_\mathrm{h}(\cdot)$ is the Huber loss function and $\sigma_{\mathrm{r}}$ is the standard deviation of the reprojection errors.
The optimized scale parameter $s$ is estimated as follows:
\begin{align}
    s = \argmin_{s \: \in \: \mathbb{R}, \{ {\bf X}_{l} \}, {\bf K_{\rm f}}} \mathcal{L} \left( s, \big\{ {\bf X}_{l} \big\}, {\bf K_{\rm f}}; \big\{ {\bf T}_{\rm v}^{(i)} \big\}, {\bf T}_{\rm s} \right)
    .
    \label{eq:cost_function_of_ba}
\end{align}
Equation (\ref{eq:cost_function_of_ba}) is a non-convex optimization problem.
Thus, it should be solved using iterative methods such as the Levenberg--Marquardt algorithm, for which an initial value is acquired by Equation (\ref{eq:scale_estimation_equation}) of Algorithm (1) or (2).
See the details of the derivation above in Section 1 of the supplementary material paper.

\section{Synthetic image experiments}
\label{sec:synthetic_image_experiments}

In Section \ref{sec:scale_estimation}, we described the two approaches of resolving scale ambiguity, with differences in the placement of the scale parameter $s$.
In this section, we investigate, via simulation, the effect of noise given to feature points on scale estimation accuracy when varying the baseline length between the two cameras of the multi-modal stereo camera system.

The scale parameter is estimated in the synthetic environment with noise in both Algorithms (1) and (2).
Preliminary experiments in the synthetic environment show that scale parameters can be estimated correctly using the proposed method when no noise is added to the feature points.
See the details under the noise-free settings in Section 2 of the supplementary material paper.

\subsection{Experimental settings}

The procedure for the synthetic image experiments is as follows:

\begin{enumerate}
    \item Scatter 3D points ${\bf X}_{\mathit i} \in \mathbb{R}^3$ $(i=1, 2, \cdots, n_{\mathrm{p}})$ randomly in a cubic space with a side length of $D$.
    \item Arrange $n_{\mathrm{c}}$ RGB--FIR camera systems in the 3D space randomly.
    More concretely, a constant relative transformation of an RGB--FIR camera system ${\bf T}_{\rm s}$ is given, and the absolute camera poses of the RGB cameras ${\bf T}_{\rm v}^{(k)}$ $(k=1, 2, \cdots, n_{\mathrm{c}})$ are set randomly. Then, the absolute camera poses of the FIR cameras ${\bf T}_{\rm f}^{(k)}$ $(k=1, 2, \cdots, n_{\mathrm{c}})$ are computed by ${\bf T}_{\rm f}^{(k)} = {\bf T}_{\rm s} {\bf T}_{\rm v}^{(k)}$.
    \item For all $k=1, 2, \cdots, n_{\mathrm{c}}$, reproject the 3D points ${\bf X}_{1}$, ${\bf X}_{2}$, $\cdots$, ${\bf X}_{\mathit n_{\mathrm{p}}}$ to the $k^{\rm{th}}$ FIR camera using ${\bf T}_{\rm f}^{(k)}$. Then, determine the normalized image points ${\bf p}_{\mathit i}^{\mathit (k)}$ $(i=1, 2, \cdots, n_{\mathrm{p}})$ using Equation (\ref{eq:convert_world_to_normalized_3D_point}).
    Gaussian noise with a standard deviation $\sigma_{\mathrm{n}} \geq 0$ can be added to all of the reprojected points.
    \item Estimate the scale parameter $s$, using both Algorithms (1) and (2) with outlier rejection based on Equation (\ref{eq:epipolar_scalar_residual}).
    Note that the true value of the scale parameter is $1.0$ because the RGB camera positions are not scaled.
\end{enumerate}
In this paper, we define $n_{\mathrm{p}} = 1000$, $D = 2000$ and $n_{\mathrm{c}} = 100$.
In addition, the relative pose ${\bf T}_{\rm s}$ between the two cameras of the camera system is set as
\begin{align}
    {\bf T}_{\rm s}
    =
    \begin{bmatrix}
        {\bf R}_{\rm s} & \; {\bf t}_{\rm s} \\
        {\bm 0}^{\mathsf T} & \; 1
    \end{bmatrix}
    \;\;
    \mathrm{with}
    \;\;
    {\bf R}_{\rm s} = {\bf I}
    \;\;
    \mathrm{and}
    \;\;
    {\bf t}_{\rm s}
    =
    \begin{bmatrix}
        d & \; 0 & \; 0
    \end{bmatrix}^\mathsf{T}
    ,
\end{align}
where $d > 0$ is the distance between the two cameras of the RGB--FIR camera system.
$d$ and $\sigma_{\mathrm{n}}$ are set depending on the simulation.

\subsection{Effects of feature point detection error}

\begin{figure}[t]
    \centering
    \begin{minipage}{0.495\hsize}
        \centering
        \includegraphics[width=6.0cm]{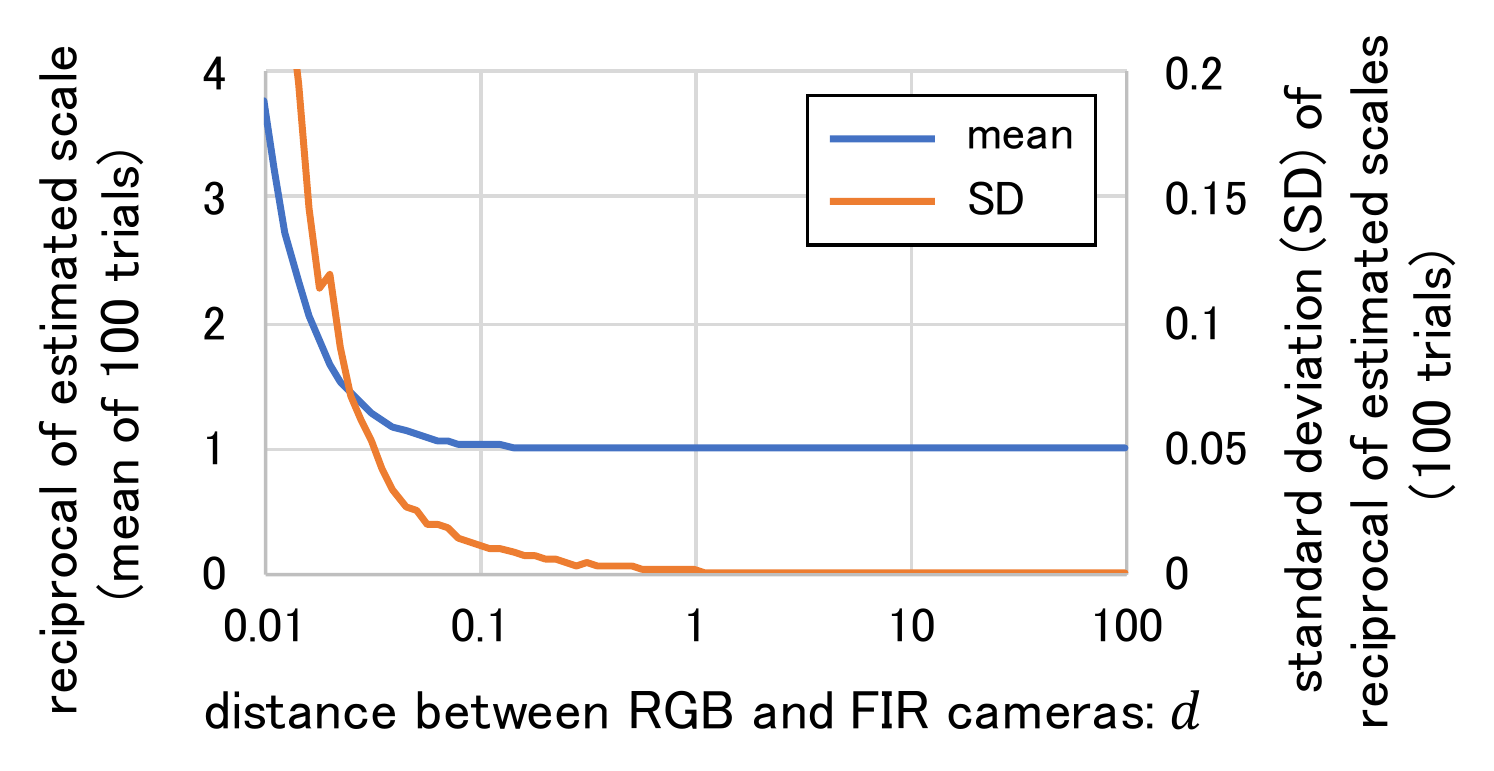}
        \subcaption{Algorithm (1)}
        \label{fig:distance_vs_scale_algo1_scale_1_noise_0_001}
    \end{minipage}
    \begin{minipage}{0.495\hsize}
        \centering
        \includegraphics[width=6.0cm]{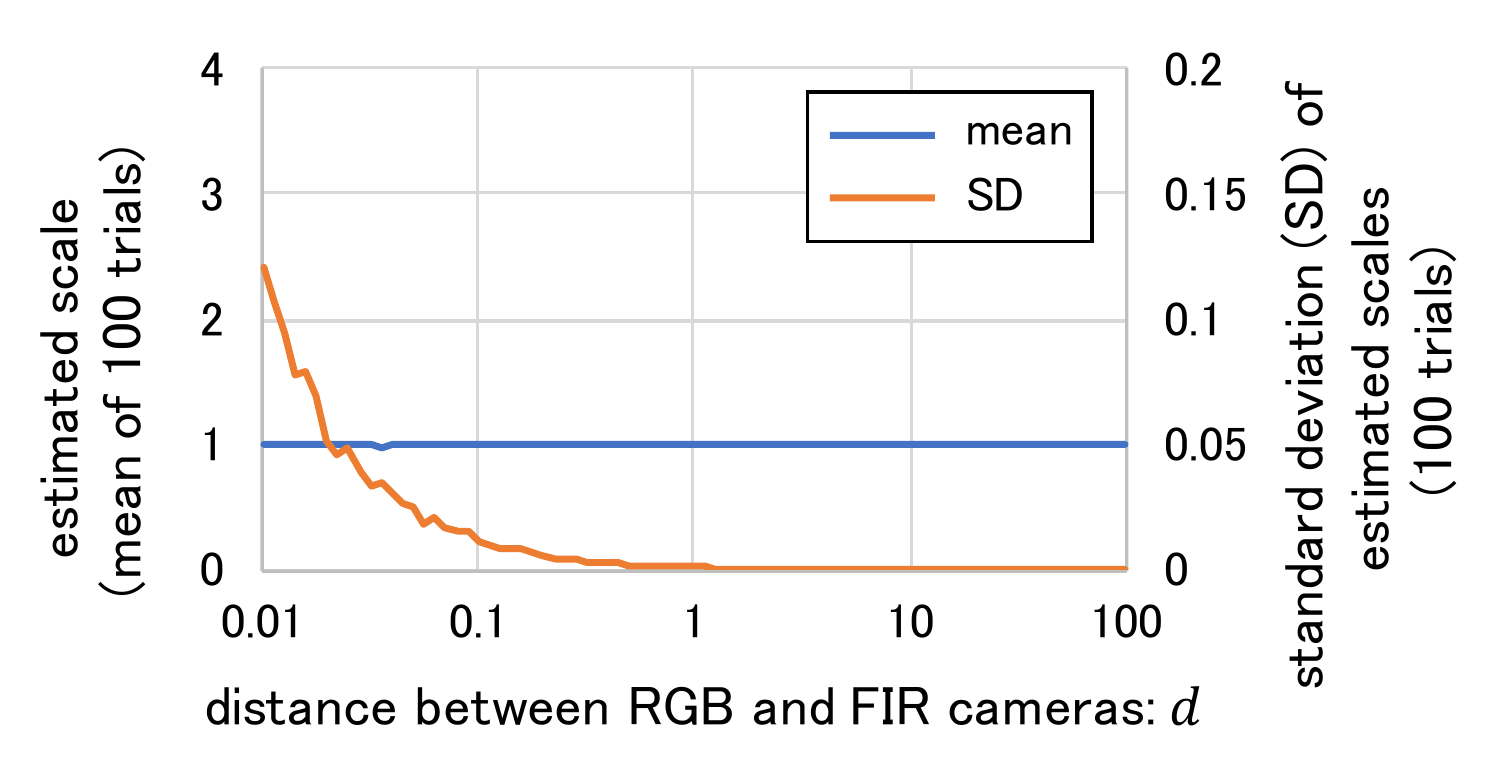}
        \subcaption{Algorithm (2)}
        \label{fig:distance_vs_scale_algo2_scale_1_noise_0_001}
    \end{minipage}
    \caption{Mean (left vertical axes) and standard deviation (right vertical axes) of a hundred estimated scales under feature point noise of $\sigma_{\mathrm{n}} = 0.001$.
    Both horizontal axes represent the baseline length $d$ between the two cameras of the RGB--FIR camera system, which is varied in the range of $[10^{-2}, 10^2]$. Note that the true value of the scale $s_{\rm true} = 1.0$ here. The accuracy and stability of the estimated scales are different between (a) and (b), especially for $d < 0.1$.}
    \label{fig:distance_vs_scale_scale_1_noise_0_001}
\end{figure}

We consider the effect of noise given to feature points on scale estimation accuracy when varying a baseline length of the stereo camera system.
Setting $\sigma_{\mathrm{n}} = 0.001$, we estimate scale parameters $s$ 100 times and compute a mean and a standard deviation (SD) of $1/s$ \big(in Algorithm (1)\big) or $s$ \big(in Algorithm (2)\big), with respect to each of the various baseline lengths $d$ between the two cameras of the camera system.
Fig.\,\ref{fig:distance_vs_scale_scale_1_noise_0_001} shows the relationship between $d$, the means and the SDs of the estimated scales for both Algorithms (1) and (2).

In Fig.\,\ref{fig:distance_vs_scale_algo2_scale_1_noise_0_001}, the scale parameters are stably estimated in the region where $d$ is relatively large ($0.1 < d$) because the means are $s = s_{\rm true} = 1.0$ and the SDs converge to $0.0$.
On the contrary, in the region where $d$ is relatively small ($d < 0.1$), the SD increases as $d$ decreases but the means maintain the correct value of $s_{\rm true} = 1.0$.
Meanwhile, in Fig.\,\ref{fig:distance_vs_scale_algo1_scale_1_noise_0_001} the means of the scale parameters are less accurate than the ones in Fig.\,\ref{fig:distance_vs_scale_algo2_scale_1_noise_0_001} in the region where $d$ is relatively small ($d < 0.1$).
In addition, the SDs in Fig.\,\ref{fig:distance_vs_scale_algo1_scale_1_noise_0_001} are larger than the ones in Fig.\,\ref{fig:distance_vs_scale_algo2_scale_1_noise_0_001}.

Hence, it is concluded that the estimated scales obtained by Algorithm (2) are more accurate and stable than the ones obtained by Algorithm (1).
Additionally, the baseline length between the two cameras of a multi-modal stereo camera system should be as long as possible for scale estimation.

\section{Real image experiments}
\label{sec:real_image_experiments}

\subsection{Evaluation method}
\label{subsec:evaluation_method}

\begin{figure}[t]
    \begin{minipage}[b]{0.63\hsize}
        \centering
        \includegraphics[height=2.0cm]{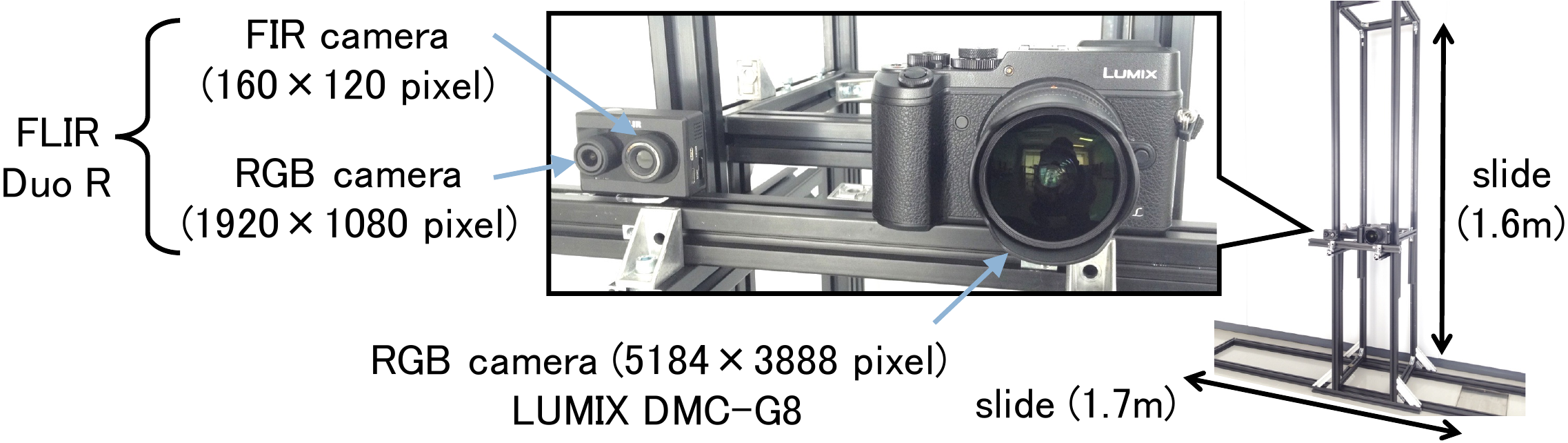}
        \subcaption{Camera system and its mount}
        \label{fig:camera_system_and_mount}
    \end{minipage}
    \begin{minipage}[b]{0.34\hsize}
        \centering
        \includegraphics[height=2.0cm]{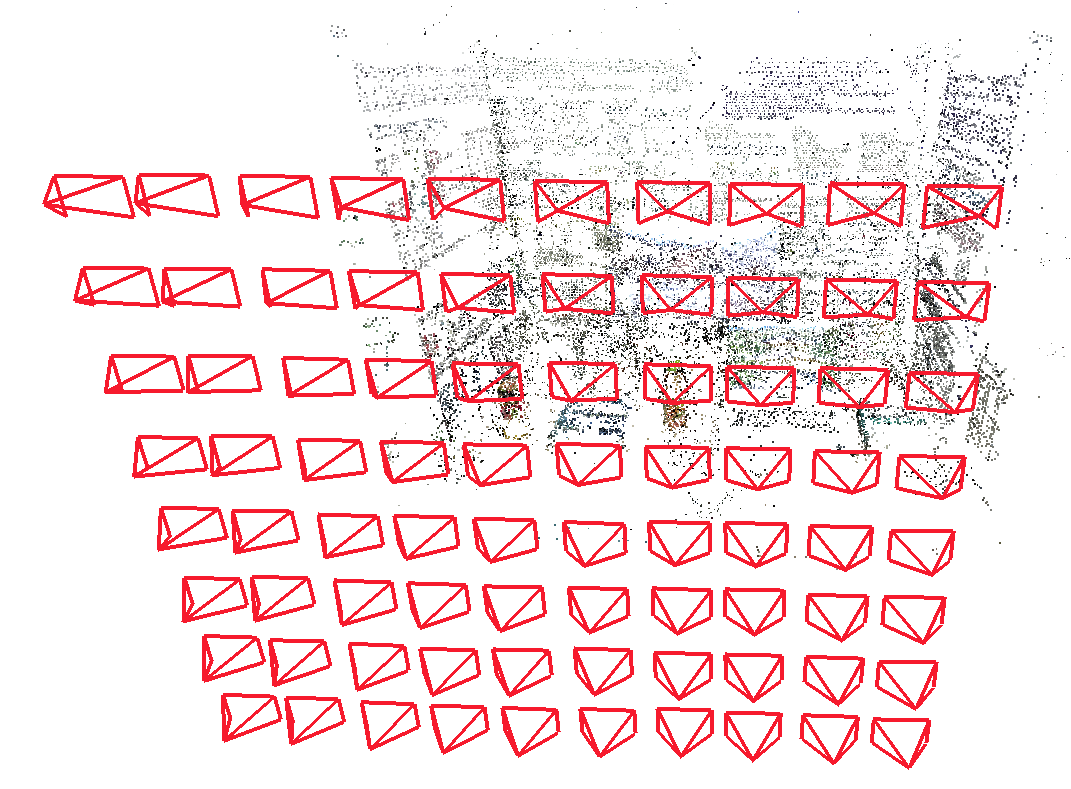}
        \subcaption{Grid-aligned cameras}
        \label{fig:grid_aligned_camera_poses}
    \end{minipage}
    \caption{\textbf{(a)} Camera system and its mount used in our experiment. The camera mount is used to capture images along the grid-aligned viewpoints. The stage, to which the camera system is fixed, can be moved in both vertical and horizontal directions.
    \textbf{(b)} Grid-aligned camera poses estimated by SfM. The images are captured using (a).}
\end{figure}

We apply the proposed method to the experimental environment to verify that the method is capable of estimating the absolute scales of outputs from monocular SfM which uses a multi-modal stereo camera.
For this verification, we need to prepare results of monocular SfM in which the actual distances between the cameras are already known.
Therefore in this experiment, the multi-modal stereo camera system is fixed to the stage on the camera mount as shown in Fig.\,\ref{fig:camera_system_and_mount}, and we capture RGB and FIR images while moving the camera system on a grid of $100\mathrm{[mm]}$ intervals.
The stage of the camera mount, where the camera system is fixed, can be moved in both vertical and horizontal directions.
Fig.\,\ref{fig:grid_aligned_camera_poses} shows an example of grid-aligned camera poses estimated by SfM, whose images are captured using the camera mount shown in Fig.\,\ref{fig:camera_system_and_mount}.

Let $d^{(ij)}$ be the actual distance between the two RGB cameras $({\rm C}_{\rm v}^{(i)}, {\rm C}_{\rm v}^{(j)})$ and $L^{(ij)}$ be the distance between $({\rm C}_{\rm v}^{(i)}, {\rm C}_{\rm v}^{(j)})$ in the result of the monocular SfM, which has scale ambiguity.
The estimated actual distance $\hat{d}^{(ij)}$ is computed by
\begin{align}
    \hat{d}^{(ij)} = s \cdot L^{(ij)}
    \textrm{\big(in Algorithm (1)\big)}
    \;
    \textrm{and}
    \;\:
    \hat{d}^{(ij)} = L^{(ij)} \big/ s
    \;
    \textrm{\big(in Algorithm (2)\big)}
    ,
\end{align}
where $s$ is the scale parameter in Equation (\ref{eq:rgb_rgb_extrinsic_with_s}) and Equation (\ref{eq:alternative_extrinsics}), respectively.
Additionally, the relative error $\epsilon^{(ij)}$ of $\hat{d}^{(ij)}$ can be defined as
\begin{align}
    \epsilon^{(ij)}
    =
    \frac{\hat{d}^{(ij)} - d^{(ij)}}{d^{(ij)}} \times 100 [\%].
\end{align}

The RGB--FIR camera system used in our experiment is shown in Fig.\,\ref{fig:camera_system_and_mount}.
The RGB camera in the camera system is a LUMIX DMC--G8 (Panasonic Corp.) or the RGB camera part of a FLIR Duo R (FLIR Systems, Inc.), depending on the experimental setting of the baseline length.
The FIR camera is the FIR camera part of the FLIR Duo R.

The procedure for the experiment is as follows:
\begin{enumerate}
    \item Capture the RGB and FIR image pairs using the camera system and its mount shown in Fig.\,\ref{fig:camera_system_and_mount}. Additionally, some supplementary RGB and FIR images are added to stabilize the process of monocular SfM and scale estimation.
    \item Perform a process of monocular SfM using the captured RGB images.
    \item Compute feature point matches of the FIR images using SIFT~\cite{lowe2004distinctive} descriptor, whose outliers are rejected via RANdom SAmple Consensus (RANSAC) based on a five-point algorithm~\cite{nister2004efficient,stewenius2006recent}.
    \item Estimate the scale parameter by Algorithms (1) and (2).
    \item Compute a mean of $\epsilon^{(ij)}$ with all the combinations, which is defined as
\begin{align}
    \overline{\epsilon}
    =
    \frac{1}{N (N - 1) / 2}
    \sum_{i < j} \epsilon^{(ij)},
\end{align}
          where $N$ is the number of RGB images taken in a grid.
\end{enumerate}

When detecting and describing feature points, FIR images are converted to gray-scaled images.
FLIR Duo R outputs FIR images whose pixels contain values of radiation temperature.
To convert them to gray-scaled images, a mean $\mu$ and a standard deviation $\sigma_{\mathrm{p}}$ of pixels for each image are computed, and then pixel values with a range of $[\mu - 2\sigma_{\mathrm{p}}, \mu + 2\sigma_{\mathrm{p}}]$ are mapped to $[0, 2^8 - 1]$.

To confirm the effect of the difference in baseline lengths between the RGB and FIR cameras, datasets of RGB and FIR images are taken with each of the four baseline lengths of the camera system: $273\mathrm{[mm]}$, $192\mathrm{[mm]}$, $113\mathrm{[mm]}$ and $26\mathrm{[mm]}$.
The systems with the first, second, and third baseline lengths use the LUMIX DMC--G8 as the RGB camera.
The system with $26\mathrm{[mm]}$ uses the RGB camera equipped on the FLIR Duo R.
Considering the randomness of RANSAC, for each of the four baseline lengths, the scale estimation and computation of $\overline{\epsilon}$ are performed 100 times.
Then, a mean and a standard deviation of $| \overline{\epsilon} |$ are calculated.

Also, pre-calibration of an RGB--FIR stereo camera system is needed to perform the proposed scale estimation procedure.
Thus, we adopt the stereo calibration method in which a planar pattern such as a chessboard is used~\cite{a-flexible-new-technique-for-camera-calibration}.
See the details in Section 3 of the supplementary material paper.

\subsection{Evaluation with a real scene}
\label{subsec:evaluation_with_the_real_scene}

\begin{figure}[t]
    \centering
    \begin{minipage}[b]{0.26\hsize}
        \centering
        \includegraphics[width=2.5cm]{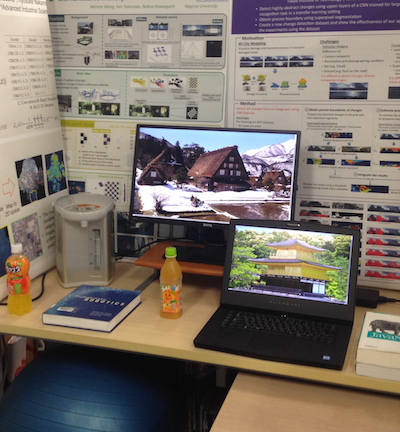}
        \subcaption{Target scene}
        \label{fig:evaluation_target_scene}
    \end{minipage}
    \begin{minipage}[b]{0.36\hsize}
        \centering
        \includegraphics[width=4.2cm]{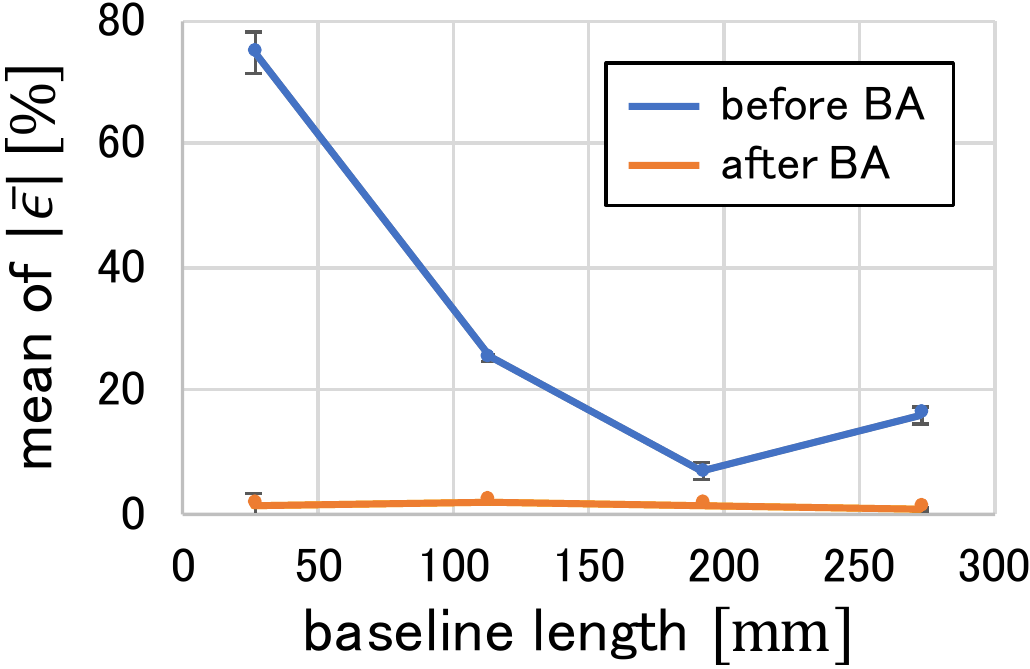}
        \subcaption{Algorithm (1)}
        \label{fig:algo_1_real_image_experiment_with_proposed}
    \end{minipage}
    \begin{minipage}[b]{0.36\hsize}
        \centering
        \includegraphics[width=4.2cm]{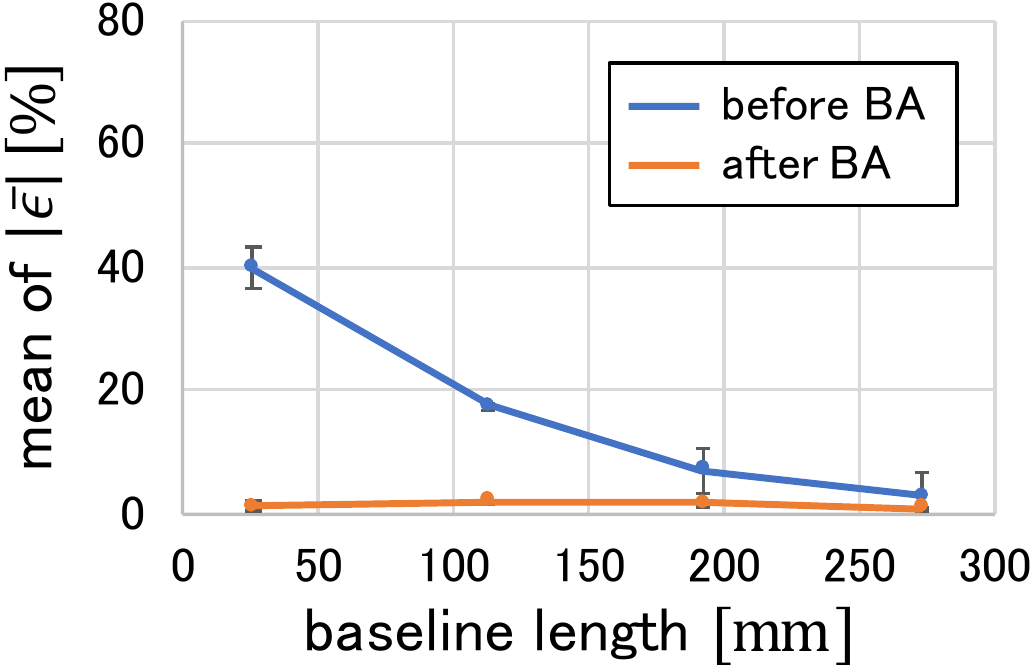}
        \subcaption{Algorithm (2)}
        \label{fig:algo_2_real_image_experiment_with_proposed}
    \end{minipage}
    \caption{\textbf{(a)} Target scene of evaluation. \textbf{(b)}, \textbf{(c)} The means of $| \overline{\epsilon} |$ with 100 trials under various baseline setups ($26, 113, 192$ and $273[\mathrm{mm}]$), in both Algorithms (1) and (2).
    The error bars indicate the range of $\pm 1 \sigma$ of $| \overline{\epsilon} |$.
    $\sigma$ is a standard deviation of $| \overline{\epsilon} |$ with 100 trials.
    It is found that the means of $| \overline{\epsilon} |$ before BA decrease as the baseline length becomes larger in both (b) and (c).
    Additionally, the means of $| \overline{\epsilon} |$ in (b) are larger the ones in (c).
    On the contrary, after BA, the means of $| \overline{\epsilon} |$ approach nearly zero in both (b) and (c).}
    \label{fig:real_image_experiment_with_proposed}
\end{figure}

The experimental environment used in the evaluation is shown in Fig.\,\ref{fig:evaluation_target_scene}.
The grid pattern along which the camera system is moved has 8 vertical $\times$ 10 horizontal grids.
Thus, there are 80 RGB camera poses used for the evaluation.
Additionally, 50 supplementary pairs of RGB and FIR images are included to stabilize the process of monocular SfM and scale estimation.
Considering the randomness of RANSAC, we show the means and standard deviations of $| \overline{\epsilon} |$ with 100 trials of scale estimation.
Figs.\,\ref{fig:algo_1_real_image_experiment_with_proposed} and \ref{fig:algo_2_real_image_experiment_with_proposed} show the results when using Algorithms (1) and (2), respectively.

In both Figs.\,\ref{fig:algo_1_real_image_experiment_with_proposed} and \ref{fig:algo_2_real_image_experiment_with_proposed}, the means of $| \overline{\epsilon} |$ before BA decrease as the baseline length becomes larger.
Additionally, the mean values in Fig.\,\ref{fig:algo_1_real_image_experiment_with_proposed} are larger than the ones in Fig.\,\ref{fig:algo_2_real_image_experiment_with_proposed} across the whole range of baseline length.
Those results denote the same pattern as the experiments in the synthetic environment in Section \ref{sec:synthetic_image_experiments}.
Consequently, without BA, it is evident that the smaller error of scale estimation occurs when using the camera system with the longer baseline as indicated by the simulation in Section \ref{sec:synthetic_image_experiments}.
In addition, the difference in numerical stability of the proposed method occurs in experiments with both synthetic and real images.

On the contrary, after BA, the means of $| \overline{\epsilon} |$ approach nearly zero in both Figs.\,\ref{fig:algo_1_real_image_experiment_with_proposed} and \ref{fig:algo_2_real_image_experiment_with_proposed}, even though large error occurred before BA.
Especially, at the $26[\mathrm{mm}]$ baseline length in Fig.\,\ref{fig:algo_1_real_image_experiment_with_proposed}, the mean of $| \overline{\epsilon} |$ after BA is $1.64[\%]$ whereas it is $74.9[\%]$ before BA.
Additionally, at the $273[\mathrm{mm}]$ baseline length after BA, high accuracy of the scale estimation is achieved as the means of $| \overline{\epsilon} |$ are $0.832[\%]$ under Algorithm (1) and $0.876[\%]$ under Algorithm (2).
The SDs also decrease after BA compared to the ones before BA.
Summarizing the above, we conclude that scale parameters estimated by both Algorithms (1) and (2) are suitable for an initial value of BA as well as that our BA effectively refines the scale parameters with respect to the accuracy and variance.

\subsection{Comparison with the existing method}

\begin{figure}[t]
    \centering
    \begin{minipage}[t]{0.54\hsize}
        \centering
        \includegraphics[height=2.8cm]{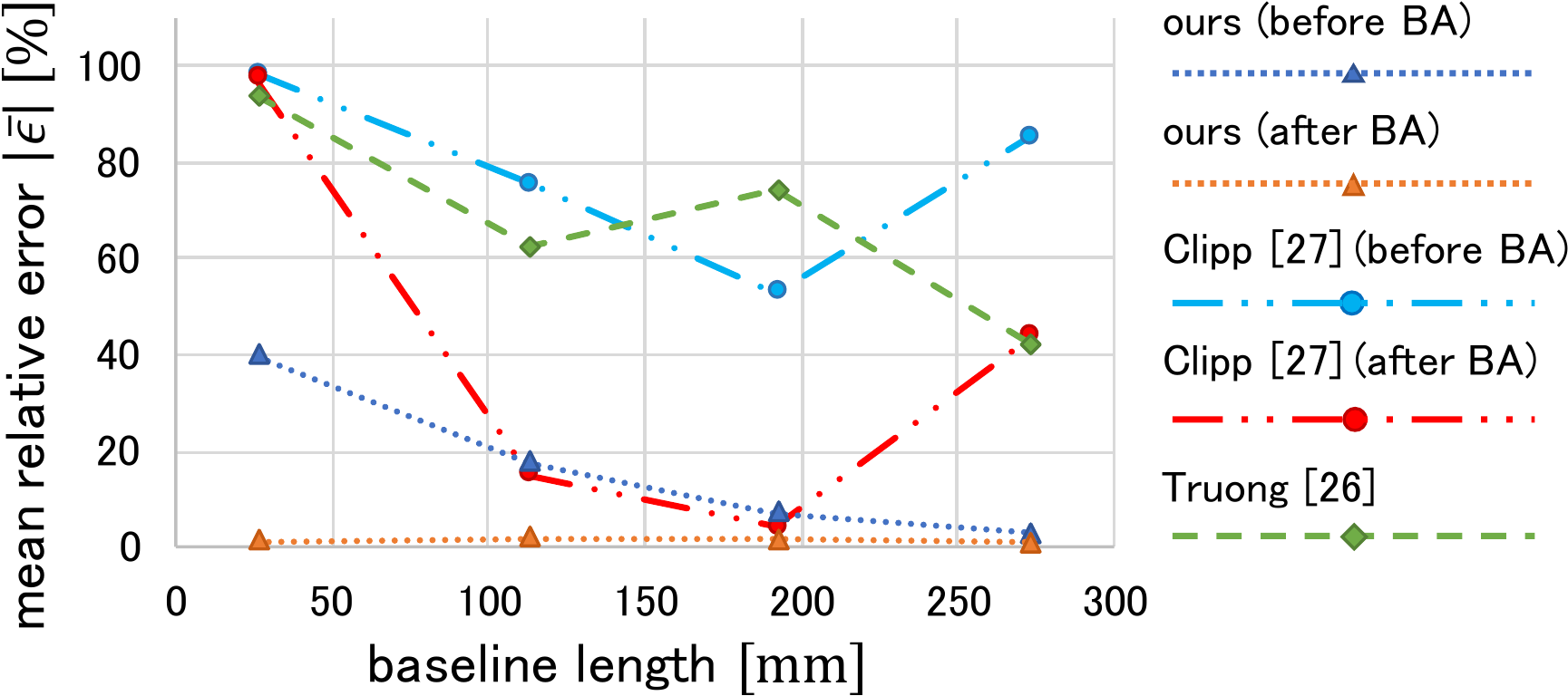}
        \caption{The mean relative errors $| \overline{\epsilon} |$ acquired by Algorithm (2) of the proposed method (\textit{ours}), \cite{Truong_2017_ICCV} and \cite{clipp2008robust}.
        \textit{Ours} and \cite{clipp2008robust} have randomness caused by RANSAC.
        Thus, we show the means of $| \overline{\epsilon} |$ with 100 trials for them.}
        \label{fig:real_image_experiment_comparison}
    \end{minipage}
    \begin{minipage}[t]{0.43\hsize}
        \centering
        \includegraphics[height=2.8cm]{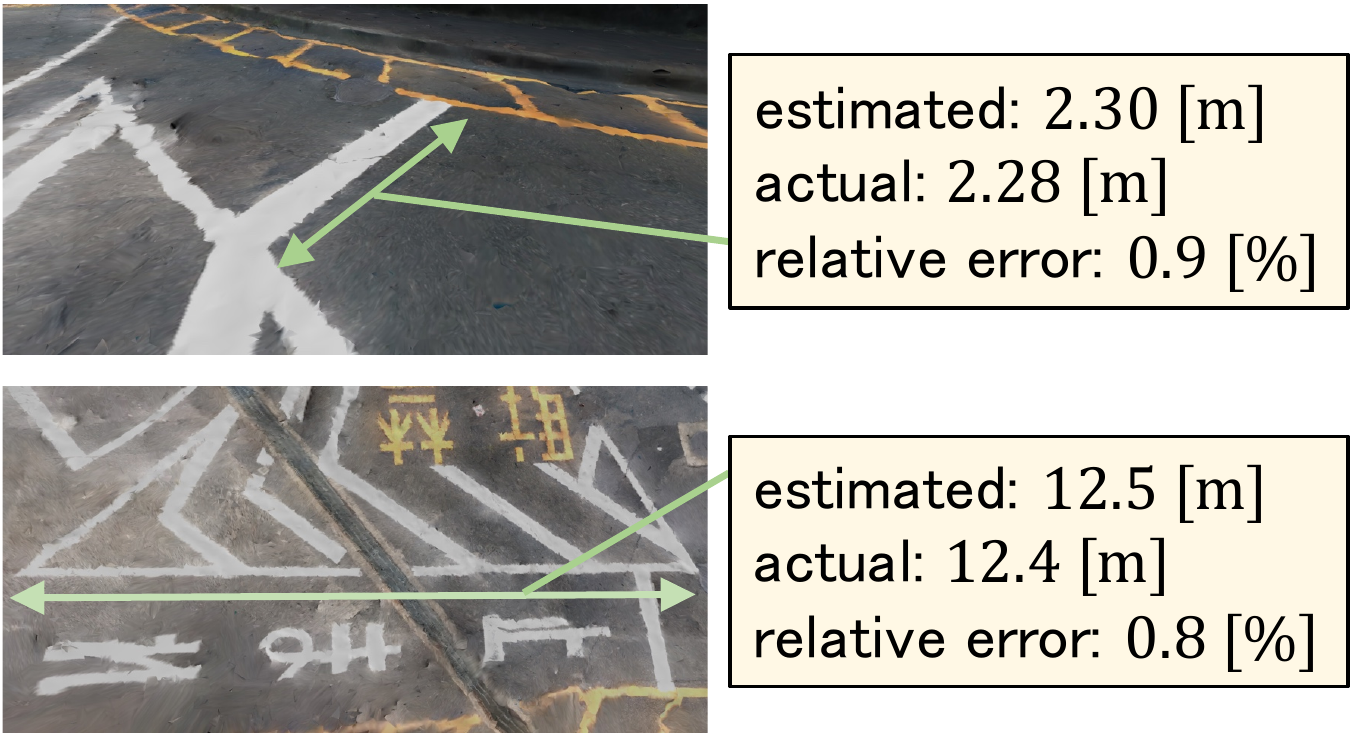}
        \caption{Evaluation of the practical result with road surface markings in the scene of Fig.\,\ref{fig:temporal_thermal_modeling_rgb}.
        High accuracy of the scale estimation is achieved as the relative errors are under $1.0[\%]$.}
        \label{fig:practical_evaluation}
    \end{minipage}
\end{figure}

As mentioned in Section \ref{subsec:scale_estimation_for_monocular_sfm}, we compare the proposed scale estimation method with the ones by Truong~\etal~\cite{Truong_2017_ICCV} and by Clipp~\etal~\cite{clipp2008robust}.
We apply the two methods of \cite{Truong_2017_ICCV} and \cite{clipp2008robust} to the RGB--FIR image datasets used in Section \ref{subsec:evaluation_with_the_real_scene}, then evaluate the estimated scale parameter by calculating $\overline{\epsilon}$ accordingly.
Fig.\,\ref{fig:real_image_experiment_comparison} shows the comparison of the accuracies of the scale parameters estimated by Algorithm (2) of the proposed method, \cite{Truong_2017_ICCV} and \cite{clipp2008robust}.
The results of the proposed method and \cite{clipp2008robust} present the means of $| \overline{\epsilon} |$ with 100 trials both before and after BA.
In result by \cite{Truong_2017_ICCV}, we adopt $s_t$ computed by Equation (6) in the paper of \cite{Truong_2017_ICCV} as the scale parameter $s$.

As shown in Fig.\,\ref{fig:real_image_experiment_comparison}, the $| \overline{\epsilon} |$ by \cite{Truong_2017_ICCV} and \cite{clipp2008robust} are much larger than the means of $| \overline{\epsilon} |$ by the proposed method throughout the whole range of baseline length.
As for \cite{Truong_2017_ICCV}, the low accuracy mainly results from the erroneous 3D points reconstructed via SfM which uses only the FIR images.
On the other hand, unlike our method, the method by \cite{clipp2008robust} cannot deal with the epipolar residuals of multiple FIR image pairs.
Thus, before BA, the means of $| \overline{\epsilon} |$ by \cite{Truong_2017_ICCV} and \cite{clipp2008robust} are much larger than the ones by the proposed method.
Additionally, the BA in \cite{clipp2008robust} does not optimize a scale parameter but rather rotations and translations.
Thus, after BA, coupled with the poor initial estimation by \cite{clipp2008robust}, the BA is unstable as shown in Fig.\,\ref{fig:real_image_experiment_comparison}.

\subsection{Practical examples}

\begin{figure}[t]
    \begin{minipage}[t]{0.315\hsize}
        \centering
        \includegraphics[width=3.7cm]{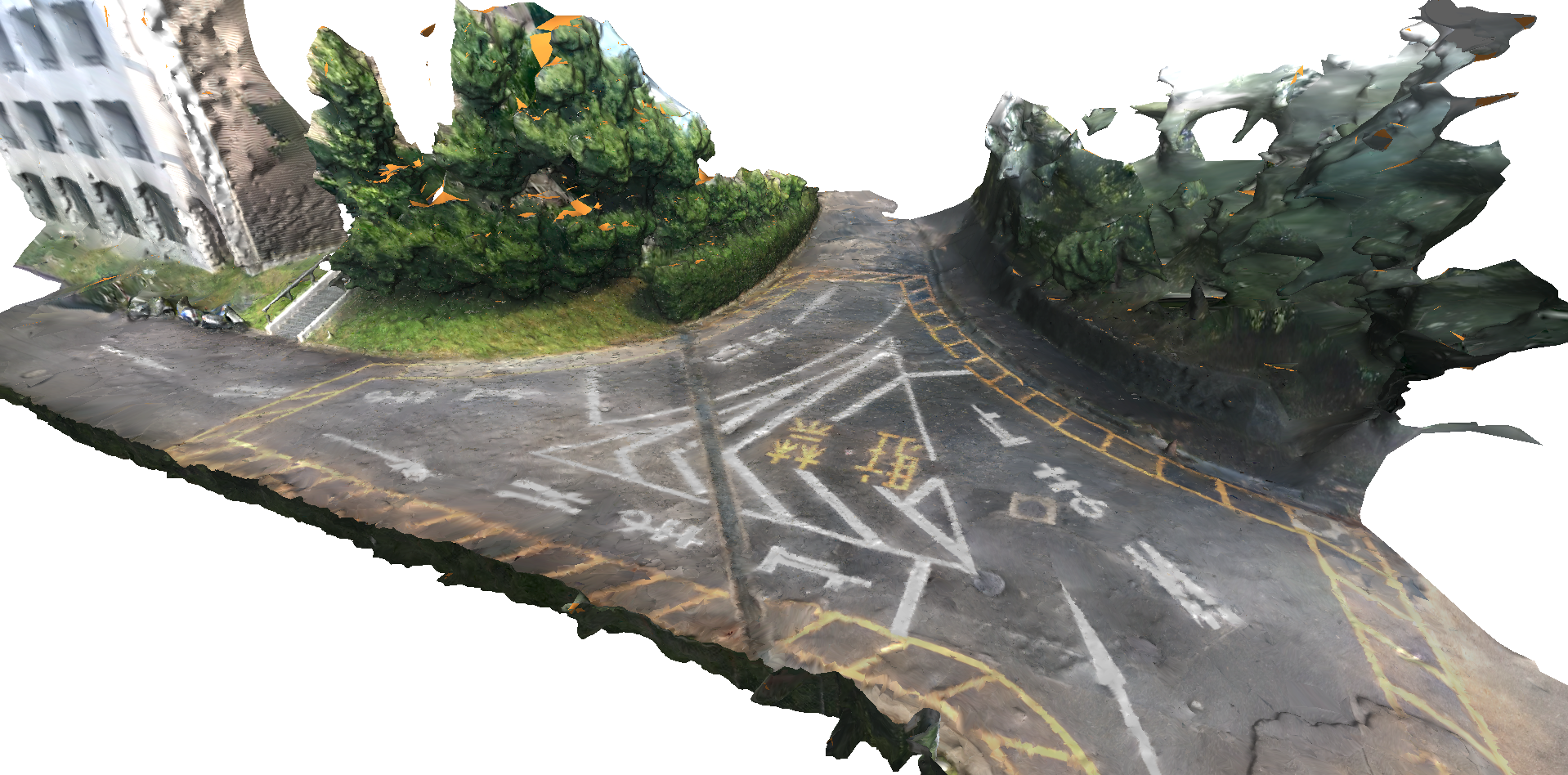}
        \subcaption{3D mesh model}
        \label{fig:temporal_thermal_modeling_rgb}
    \end{minipage}
    \begin{minipage}[t]{0.315\hsize}
        \centering
        \includegraphics[width=3.7cm]{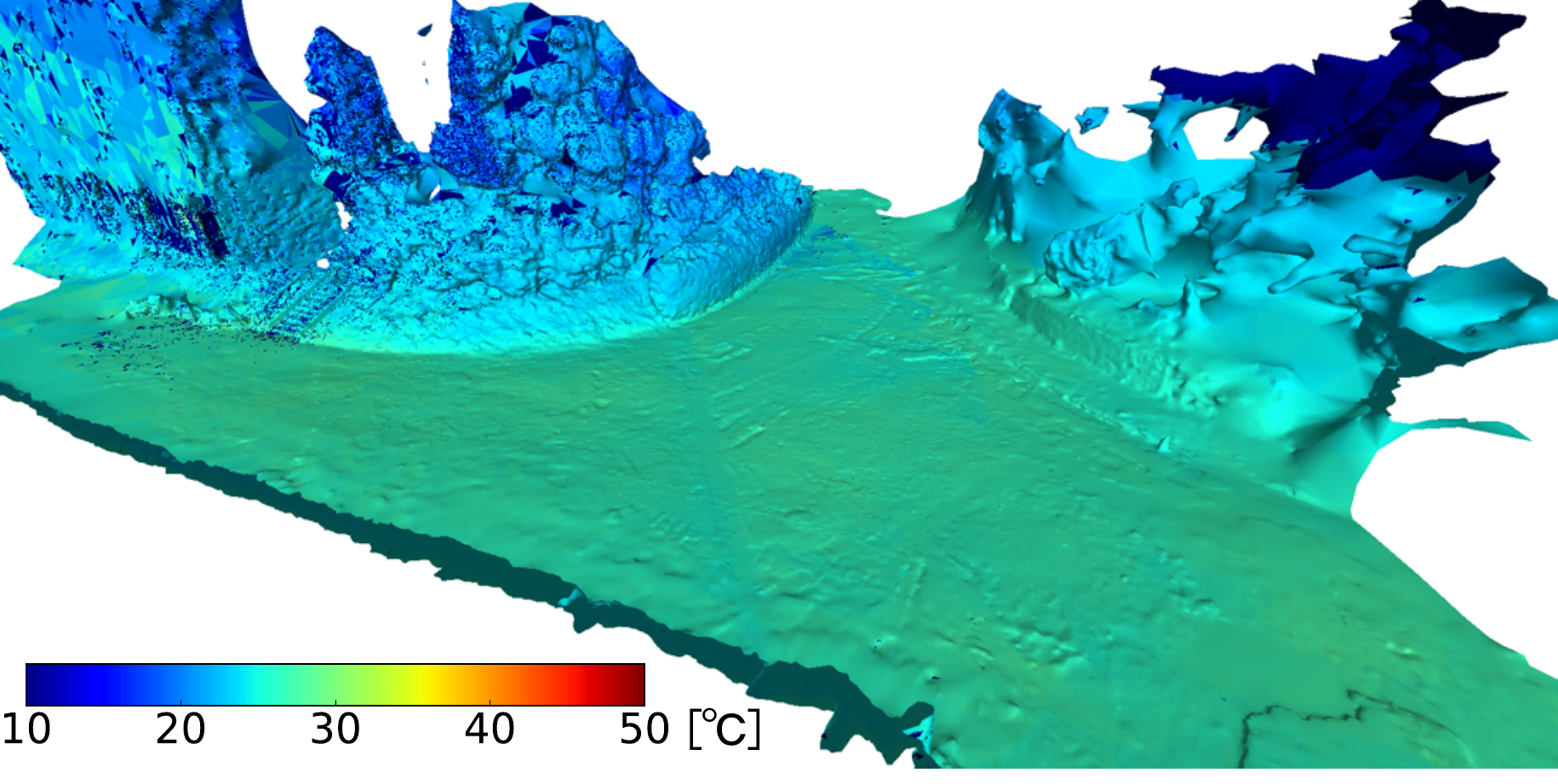}
        \subcaption{on a rainy day}
        \label{fig:temporal_thermal_modeling_rainy}
    \end{minipage}
    \begin{minipage}[t]{0.315\hsize}
        \centering
        \includegraphics[width=3.7cm]{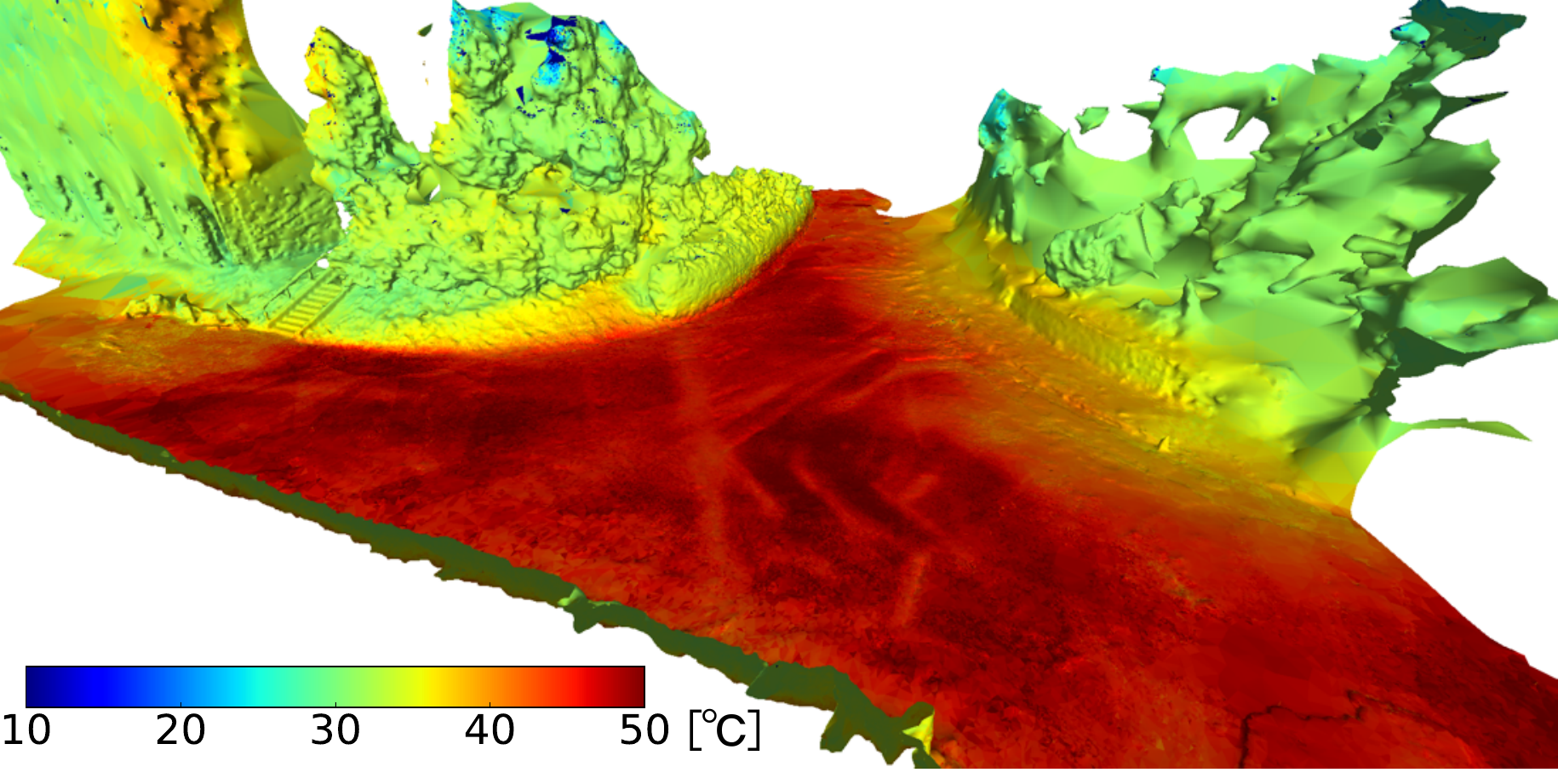}
        \subcaption{on a sunny day}
        \label{fig:temporal_thermal_modeling_sunny}
    \end{minipage}
    \caption{Examples of temporal thermal 3D modeling. The 3D mesh model reconstructed from RGB images is shown in (a). (b) and (c) show the thermal 3D reconstructions on a rainy day and on a sunny day, respectively.}
    \label{fig:temporal_thermal_modeling}
\end{figure}

Fig.\,\ref{fig:temporal_thermal_modeling} presents temporal thermal 3D mappings as a practical example of thermal 3D reconstruction.
RGB and FIR images are captured by a smartphone-based RGB--FIR camera system, composed of a FLIR One (FLIR Systems, Inc.) and a smartphone.
The baseline length of the camera system is $154[\mathrm{mm}]$.

A 3D mesh model shown in Fig.\,\ref{fig:temporal_thermal_modeling_rgb} is reconstructed from the RGB images using monocular SfM and MVS, and is then resized to the absolute scale estimated by the proposed method.
The thermal 3D models shown in Figs.\,\ref{fig:temporal_thermal_modeling_rainy} and \ref{fig:temporal_thermal_modeling_sunny} are built by reprojecting FIR images to the 3D mesh model on a sunny day and on a rainy day, respectively.
The thermal information is reprojected well as shown in Figs.\,\ref{fig:temporal_thermal_modeling_rainy} and \ref{fig:temporal_thermal_modeling_sunny}.
In addition, as shown in Fig.\,\ref{fig:practical_evaluation}, we measure the size of road surface markings in the 3D model (\textit{estimated}) and in the real world (\textit{actual}), as an evaluation of the estimated scales in practical scenes.
The relative errors of the estimated size are approximately $0.8[\%]$ in the scene in Fig.\,\ref{fig:temporal_thermal_modeling}.
See the additional results in Section 4 of the supplementary material paper.

\section{Conclusion}
\label{sec:conclusion}

In this paper, we have shown a novel method of estimating the scale parameter of monocular SfM for a multi-modal stereo camera system, which is composed of different spectral cameras (\eg RGB and FIR) in a stereo camera setup.
Owing to the difficulty of matching feature points directly between RGB and FIR images, we have leveraged a constant extrinsic parameter of the stereo setup and a small number of feature correspondences between the same modal images.
Two types of formulae for scale parameter estimation, both of which are based on the epipolar constraint, were proposed in this paper.
We have also verified the difference in scale estimation accuracy and stability between the two formulae in the synthetic and real image experiments.
The cause for the difference in scale estimation stability requires further investigation.

Additionally, we have demonstrated a scale estimation of monocular SfM under the experimental environment using an RGB--FIR stereo camera, and we have verified its accuracy both before and after BA.
The consequence shows that the proposed method can estimate an appropriate scale parameter and its accuracy depends on the baseline length between RGB and FIR cameras of a stereo camera system.
Moreover, we have presented the thermal 3D modeling as an application of the proposed scale estimation method.

These results suggest that the proposed method is applicable to the construction of thermal 3D mappings using payload-limited vehicles, such as UAVs, on which an RGB--FIR camera system is mounted.
Therefore, we conclude that the proposed method is suitable for scale estimation of monocular SfM.

\leavevmode\newline
\noindent
{\bf Acknowledgements.}
This research is supported by the Hori Sciences \& Arts Foundation, the New Energy and Industrial Technology Development Organization (NEDO) and JSPS KAKENHI Grant Number 18K18071.

\bibliographystyle{splncs04}
\bibliography{0672}

\begin{thebibliography}{10}
\providecommand{\url}[1]{\texttt{#1}}
\providecommand{\urlprefix}{URL }
\providecommand{\doi}[1]{https://doi.org/#1}

\bibitem{Agarwal2009}
Agarwal, S., Snavely, N., Simon, I., Seitz, S.M., Szeliski, R.: {Building Rome
  in a day}. In: International Conference on Computer Vision (ICCV). pp. 72--79
  (2009)

\bibitem{bay2006surf}
Bay, H., Tuytelaars, T., Van~Gool, L.: Surf: Speeded up robust features. In:
  European Conference on Computer Vision (ECCV). pp. 404--417 (2006)

\bibitem{BERTOZZI2007194}
Bertozzi, M., Broggi, A., Caraffi, C., Rose, M.D., Felisa, M., Vezzoni, G.:
  Pedestrian detection by means of far-infrared stereo vision. Computer Vision
  and Image Understanding  \textbf{106}(2),  194--204 (2007)

\bibitem{clipp2008robust}
Clipp, B., Kim, J.H., Frahm, J.M., Pollefeys, M., Hartley, R.: Robust 6dof
  motion estimation for non-overlapping, multi-camera systems. In: IEEE
  Workshop on Applications of Computer Vision (WACV) (2008)

\bibitem{MonoSLAM}
Davison, A.J., Reid, I.D., Molton, N.D., Stasse, O.: Monoslam: Real-time single
  camera slam. Transactions on Pattern Analysis and Machine Intelligence
  (TPAMI)  \textbf{29}(6),  1052--1067 (2007)

\bibitem{detone2017toward}
DeTone, D., Malisiewicz, T., Rabinovich, A.: Toward geometric deep slam. arXiv
  preprint arXiv:1707.07410  (2017)

\bibitem{Furukawa2010}
Furukawa, Y., Ponce, J.: {Accurate, Dense, and Robust Multi-View Stereopsis}.
  Transactions on Pattern Analysis and Machine Intelligence (TPAMI)
  \textbf{32}(8),  1362--1376 (2010)

\bibitem{HAM2013395}
Ham, Y., Golparvar-Fard, M.: An automated vision-based method for rapid 3d
  energy performance modeling of existing buildings using thermal and digital
  imagery. Advanced Engineering Informatics  \textbf{27}(3),  395--409 (2013)

\bibitem{han2015matchnet}
Han, X., Leung, T., Jia, Y., Sukthankar, R., Berg, A.C.: Matchnet: Unifying
  feature and metric learning for patch-based matching. In: Conference on
  Computer Vision and Pattern Recognition (CVPR). pp. 3279--3286 (2015)

\bibitem{Hartley2004}
Hartley, R.I., Zisserman, A.: Multiple View Geometry in Computer Vision.
  Cambridge University Press, ISBN: 0521540518, second edn. (2004)

\bibitem{iwaszczuk2017camera}
Iwaszczuk, D., Stilla, U.: Camera pose refinement by matching uncertain 3d
  building models with thermal infrared image sequences for high quality
  texture extraction. ISPRS Journal of Photogrammetry and Remote Sensing
  \textbf{132},  33--47 (2017)

\bibitem{jancosek2011multi}
Jancosek, M., Pajdla, T.: Multi-view reconstruction preserving weakly-supported
  surfaces. In: Conference on Computer Vision and Pattern Recognition (CVPR).
  pp. 3121--3128 (2011)

\bibitem{Kitt20117357}
Kitt, B.M., Rehder, J., Chambers, A.D., Schonbein, M., Lategahn, H., Singh, S.:
  Monocular visual odometry using a planar road model to solve scale ambiguity.
  In: European Conference on Mobile Robots (2011)

\bibitem{klein2007parallel}
Klein, G., Murray, D.: Parallel tracking and mapping for small ar workspaces.
  In: International Symposium on Mixed and Augmented Reality (ISMAR). pp.
  225--234 (2007)

\bibitem{lowe2004distinctive}
Lowe, D.G.: Distinctive image features from scale-invariant keypoints.
  International Journal of Computer Vision (IJCV)  \textbf{60}(2),  91--110
  (2004)

\bibitem{7676356}
M^^c3^^bcller, A.O., Kroll, A.: Generating high fidelity 3-d thermograms with a
  handheld real-time thermal imaging system. IEEE Sensors Journal
  \textbf{17}(3),  774--783 (2017)

\bibitem{newcombe2011kinectfusion}
Newcombe, R.A., Izadi, S., Hilliges, O., Molyneaux, D., Kim, D., Davison, A.J.,
  Kohi, P., Shotton, J., Hodges, S., Fitzgibbon, A.: Kinectfusion: Real-time
  dense surface mapping and tracking. In: International symposium on Mixed and
  augmented reality (ISMAR). pp. 127--136 (2011)

\bibitem{nister2004efficient}
Nist{\'e}r, D.: An efficient solution to the five-point relative pose problem.
  IEEE transactions on pattern analysis and machine intelligence
  \textbf{26}(6),  756--770 (2004)

\bibitem{Nuetzi2011}
N{\"u}tzi, G., Weiss, S., Scaramuzza, D., Siegwart, R.: Fusion of imu and
  vision for absolute scale estimation in monocular slam. Journal of
  Intelligent {\&} Robotic Systems  \textbf{61}(1),  287--299 (2011)

\bibitem{cramer2014automatic}
Oreifej, O., Cramer, J., Zakhor, A.: Automatic generation of 3d thermal maps of
  building interiors. ASHRAE transactions  \textbf{120}, ~C1 (2014)

\bibitem{Truong_2017_ICCV}
Phuc~Truong, T., Yamaguchi, M., Mori, S., Nozick, V., Saito, H.: Registration
  of rgb and thermal point clouds generated by structure from motion. In:
  International Conference on Computer Vision Workshop (ICCVW) (2017)

\bibitem{Rublee2011}
Rublee, E., Rabaud, V., Konolige, K., Bradski, G.: {ORB: an efficient
  alternative to SIFT or SURF}. In: International Conference on Computer Vision
  (ICCV). pp. 2564--2571 (2011)

\bibitem{scaramuzza2009absolute}
Scaramuzza, D., Fraundorfer, F., Pollefeys, M., Siegwart, R.: Absolute scale in
  structure from motion from a single vehicle mounted camera by exploiting
  nonholonomic constraints. In: International Conference on Computer Vision
  (ICCV). pp. 1413--1419 (2009)

\bibitem{schoenberger2016sfm}
Sch\"{o}nberger, J.L., Frahm, J.M.: Structure-from-motion revisited. In:
  Conference on Computer Vision and Pattern Recognition (CVPR). pp. 4104--4113
  (2016)

\bibitem{schoenberger2016mvs}
Sch\"{o}nberger, J.L., Zheng, E., Pollefeys, M., Frahm, J.M.: Pixelwise view
  selection for unstructured multi-view stereo. In: European Conference on
  Computer Vision (ECCV). pp. 501--518 (2016)

\bibitem{stewenius2006recent}
Stew{\'e}nius, H., Engels, C., Nist{\'e}r, D.: Recent developments on direct
  relative orientation. ISPRS Journal of Photogrammetry and Remote Sensing
  \textbf{60},  284--294 (2006)

\bibitem{THIELE2017140}
Thiele, S.T., Varley, N., James, M.R.: Thermal photogrammetric imaging: A new
  technique for monitoring dome eruptions. Journal of Volcanology and
  Geothermal Research  \textbf{337}(Supplement C),  140--145 (2017)

\bibitem{triggs1999bundle}
Triggs, B., McLauchlan, P.F., Hartley, R.I., Fitzgibbon, A.W.: Bundle
  adjustment --- a modern synthesis. In: Vision Algorithms: Theory and
  Practice. pp. 298--372 (1999)

\bibitem{vidas20133d}
Vidas, S., Moghadam, P., Bosse, M.: 3d thermal mapping of building interiors
  using an rgb-d and thermal camera. In: International Conference on Robotics
  and Automation (ICRA). pp. 2311--2318 (2013)

\bibitem{weinmann2014thermal}
Weinmann, M., Leitloff, J., Hoegner, L., Jutzi, B., Stilla, U., Hinz, S.:
  Thermal 3d mapping for object detection in dynamic scenes. ISPRS Annals of
  the Photogrammetry, Remote Sensing and Spatial Information Sciences
  \textbf{2}(1), ~53 (2014)

\bibitem{Zagoruyko_2015_CVPR}
Zagoruyko, S., Komodakis, N.: Learning to compare image patches via
  convolutional neural networks. In: Conference on Computer Vision and Pattern
  Recognition (CVPR). pp. 4353--4361 (2015)

\bibitem{a-flexible-new-technique-for-camera-calibration}
Zhang, Z.: A flexible new technique for camera calibration. IEEE Transactions
  on Pattern Analysis and Machine Intelligence (TPAMI)  \textbf{22},
  1330^^e2^^80^^93--1334 (2000)

\end{thebibliography}



\section*{Supplementary material}

# Supplementary Material for "Scale Estimation of Monocular SfM for a Multi-modal Stereo Camera"

**Contents**

## 1 Details of the derivation

This section explains the derivation process of the scale estimation formulae in detail. In Section 1.1, we describe the definitions of the vectors used in the derivation. Section 1.2 explains the differentiation of the error function for a least-squares method. Section 1.3 describes the derivation of the projection function used in the scale-oriented bundle adjustment (BA) in detail.

### 1.1 Definitions of the vectors: $\mathbf{u}_k^{(ij)}$, $\mathbf{f}^{(ij)}$ and $\mathbf{g}^{(ij)}$

In the main paper, the relative transformation matrix between the two FIR cameras, $C_f^{(i)}$ and $C_f^{(j)}$, is explained as

$$\mathbf{T}_f^{(ij)} = \mathbf{T}_s \mathbf{T}_v^{(ij)} \mathbf{T}_s^{-1} \tag{1}$$

$$= \begin{bmatrix} \mathbf{A}^{(ij)} & s \cdot \mathbf{b}^{(ij)} + \mathbf{c}^{(ij)} \\ \mathbf{0}^\mathsf{T} & 1 \end{bmatrix}. \tag{2}$$

In Algorithm (1), $\mathbf{A}^{(ij)}$, $\mathbf{b}^{(ij)}$ and $\mathbf{c}^{(ij)}$ are defined as follows:

$$\begin{cases} \mathbf{A}^{(ij)} = [\mathbf{a}_1^{(ij)} | \mathbf{a}_2^{(ij)} | \mathbf{a}_3^{(ij)}] = \mathbf{R}_s \mathbf{R}_v^{(ij)} \mathbf{R}_s^{-1} \\ \mathbf{b}^{(ij)} = \mathbf{R}_s \mathbf{t}_v^{(ij)} \\ \mathbf{c}^{(ij)} = (\mathbf{I} - \mathbf{R}_s \mathbf{R}_v^{(ij)} \mathbf{R}_s^{-1}) \mathbf{t}_s. \end{cases} \tag{3}$$

On the other hand, in Algorithm (2), they are defined as follows:

$$\begin{cases} \mathbf{A}^{(ij)} = [\mathbf{a}_1^{(ij)} | \mathbf{a}_2^{(ij)} | \mathbf{a}_3^{(ij)}] = \mathbf{R}_s \mathbf{R}_v^{(ij)} \mathbf{R}_s^{-1} \\ \mathbf{b}^{(ij)} = (\mathbf{I} - \mathbf{R}_s \mathbf{R}_v^{(ij)} \mathbf{R}_s^{-1}) \mathbf{t}_s \\ \mathbf{c}^{(ij)} = \mathbf{R}_s \mathbf{t}_v^{(ij)}. \end{cases} \tag{4}$$

Note that $\mathbf{a}_1^{(ij)}$, $\mathbf{a}_1^{(ij)}$ and $\mathbf{a}_3^{(ij)}$ are the three-dimensional column vectors.

The essential matrix $\mathbf{E}^{(ij)}$ between the two FIR cameras, $C_f^{(i)}$ and $C_f^{(j)}$, can be derived from $\mathbf{T}_f^{(ij)}$ and expressed as

$$\mathbf{E}^{(ij)} = \left[s\mathbf{b}^{(ij)} + \mathbf{c}^{(ij)}\right]_\times \mathbf{A}^{(ij)} \tag{5}$$

$$= s \cdot \left[\mathbf{b}^{(ij)} \times \mathbf{a}_1^{(ij)} \mid \mathbf{b}^{(ij)} \times \mathbf{a}_2^{(ij)} \mid \mathbf{b}^{(ij)} \times \mathbf{a}_3^{(ij)}\right]$$
$$+ \left[\mathbf{c}^{(ij)} \times \mathbf{a}_1^{(ij)} \mid \mathbf{c}^{(ij)} \times \mathbf{a}_2^{(ij)} \mid \mathbf{c}^{(ij)} \times \mathbf{a}_3^{(ij)}\right]. \tag{6}$$

The operator $[\cdot]_\times$, called a skew-symmetric matrix, is defined by

$$[\mathbf{v}]_\times = \begin{bmatrix} 0 & v_3 & -v_2 \\ -v_3 & 0 & v_1 \\ v_2 & -v_1 & 0 \end{bmatrix} \text{ with } \mathbf{v} = \begin{bmatrix} v_1 \\ v_2 \\ v_3 \end{bmatrix}. \tag{7}$$

The epipolar constraint between the two FIR images, $I_f^{(i)}$ and $I_f^{(j)}$, corresponding to the FIR cameras, $C_f^{(i)}$ and $C_f^{(j)}$, is formulated as

$$\mathbf{p}_k^{(j)^\mathsf{T}} \mathbf{E}^{(ij)} \mathbf{p}_k^{(i)} = 0, \tag{8}$$

where $\mathbf{p}_k^{(i)} = \left[x_k^{(i)}, y_k^{(i)}, 1\right]^\mathsf{T}$ and $\mathbf{p}_k^{(j)} = \left[x_k^{(j)}, y_k^{(j)}, 1\right]^\mathsf{T}$ are the $k^\text{th}$ corresponding feature points between $I_f^{(i)}$ and $I_f^{(j)}$, respectively, in the form of normalized image coordinates. A normalized image point $\mathbf{p}_k^{(i)}$ is defined as

$$\mathbf{p}_k^{(i)} = \mathbf{K}_f^{-1} \left[u_k^{(i)}, v_k^{(i)}, 1\right]^\mathsf{T} \text{ with } \mathbf{K}_f = \begin{bmatrix} f_x & 0 & c_x \\ 0 & f_y & c_y \\ 0 & 0 & 1 \end{bmatrix}, \tag{9}$$

where $\mathbf{K}_f$ is the intrinsic parameter matrix of the FIR camera. $\left[u_k^{(i)}, v_k^{(i)}\right]$ is the feature point in pixels in $I_f^{(i)}$, and is the $k^\text{th}$ corresponding feature point with $\left[u_k^{(j)}, v_k^{(j)}\right]$ in $I_f^{(j)}$.

Equation (8) can be expanded to

$$\mathbf{u}_k^{(ij)} \left(s \cdot \mathbf{f}^{(ij)} + \mathbf{g}^{(ij)}\right) = 0 \tag{10}$$

with

$$\mathbf{u}_k^{(ij)} = \left[\, x_k^{(i)} x_k^{(j)},\, x_k^{(i)} y_k^{(j)},\, x_k^{(i)},\, y_k^{(i)} x_k^{(j)},\, y_k^{(i)} y_k^{(j)},\, y_k^{(i)},\, x_k^{(j)},\, y_k^{(j)},\, 1\, \right], \tag{11}$$

$$\mathbf{f}^{(ij)} = \Big[\, \left[\mathbf{b}^{(ij)} \times \mathbf{a}_1^{(ij)}\right]_1,\, \left[\mathbf{b}^{(ij)} \times \mathbf{a}_1^{(ij)}\right]_2,\, \left[\mathbf{b}^{(ij)} \times \mathbf{a}_1^{(ij)}\right]_3,$$
$$\left[\mathbf{b}^{(ij)} \times \mathbf{a}_2^{(ij)}\right]_1,\, \left[\mathbf{b}^{(ij)} \times \mathbf{a}_2^{(ij)}\right]_2,\, \left[\mathbf{b}^{(ij)} \times \mathbf{a}_2^{(ij)}\right]_3,$$
$$\left[\mathbf{b}^{(ij)} \times \mathbf{a}_3^{(ij)}\right]_1,\, \left[\mathbf{b}^{(ij)} \times \mathbf{a}_3^{(ij)}\right]_2,\, \left[\mathbf{b}^{(ij)} \times \mathbf{a}_3^{(ij)}\right]_3 \,\Big]^\mathsf{T}, \tag{12}$$

$$\mathbf{g}^{(ij)} = \Big[\ [\mathbf{c}^{(ij)} \times \mathbf{a}_1^{(ij)}]_1 \,,\ [\mathbf{c}^{(ij)} \times \mathbf{a}_1^{(ij)}]_2 \,,\ [\mathbf{c}^{(ij)} \times \mathbf{a}_1^{(ij)}]_3 \,,$$
$$[\mathbf{c}^{(ij)} \times \mathbf{a}_2^{(ij)}]_1 \,,\ [\mathbf{c}^{(ij)} \times \mathbf{a}_2^{(ij)}]_2 \,,\ [\mathbf{c}^{(ij)} \times \mathbf{a}_2^{(ij)}]_3 \,,$$
$$[\mathbf{c}^{(ij)} \times \mathbf{a}_3^{(ij)}]_1 \,,\ [\mathbf{c}^{(ij)} \times \mathbf{a}_3^{(ij)}]_2 \,,\ [\mathbf{c}^{(ij)} \times \mathbf{a}_3^{(ij)}]_3 \ \Big]^\mathsf{T}, \quad (13)$$

where $\mathbf{u}_k^{(ij)}$ is the row vector whereas $\mathbf{f}^{(ij)}$ and $\mathbf{g}^{(ij)}$ are the column vectors. The notation of $[\mathbf{v}]_k$ represents the $k^{\text{th}}$ element of the vector $\mathbf{v}$.

## 1.2   Differentiation of the error function $J(s)$

In the main paper, the error function $J(s)$ is described by

$$J(s) = \frac{1}{2} \sum_{i,j,i\neq j} \left\| \mathbf{e}^{(ij)} \right\|^2 \quad (14)$$

$$= \frac{1}{2} \sum_{i,j,i\neq j} \left\| \mathbf{U}^{(ij)} \big(s \cdot \mathbf{f}^{(ij)} + \mathbf{g}^{(ij)}\big) \right\|^2, \quad (15)$$

where $\mathbf{e}^{(ij)}$ is the residual vector based on the epipolar constraint between the $i^{\text{th}}$ and $j^{\text{th}}$ FIR images, which are notated by $\mathrm{I}_\mathrm{f}^{(i)}$ and $\mathrm{I}_\mathrm{f}^{(j)}$, respectively. $J(s)$ can be expanded as follows:

$$J(s) = \frac{1}{2} \sum_{i,j,i\neq j} \left( \mathbf{U}^{(ij)}\big(s \cdot \mathbf{f}^{(ij)} + \mathbf{g}^{(ij)}\big) \right)^\mathsf{T} \left( \mathbf{U}^{(ij)}\big(s \cdot \mathbf{f}^{(ij)} + \mathbf{g}^{(ij)}\big) \right) \quad (16)$$

$$= \frac{1}{2} \sum_{i,j,i\neq j} \Big( s^2 \cdot \mathbf{f}^{(ij)\mathsf{T}} \mathbf{U}^{(ij)\mathsf{T}} \mathbf{U}^{(ij)} \mathbf{f}^{(ij)}$$
$$+ 2s \cdot \mathbf{f}^{(ij)\mathsf{T}} \mathbf{U}^{(ij)\mathsf{T}} \mathbf{U}^{(ij)} \mathbf{g}^{(ij)} + \mathbf{g}^{(ij)\mathsf{T}} \mathbf{U}^{(ij)\mathsf{T}} \mathbf{U}^{(ij)} \mathbf{g}^{(ij)} \Big). \quad (17)$$

As mentioned in the main paper, the scale estimation problem comes down to determining $s$, such that the error function $J(s)$ is minimized. In other words, the scale parameter $s$ is determined by solving the equation $\mathrm{d}J(s)/\mathrm{d}s = 0$ in terms of $s$. The differentiation of $J(s)$ is

$$\frac{\mathrm{d}J(s)}{\mathrm{d}s} = \frac{\mathrm{d}}{\mathrm{d}s}\left\{ \frac{1}{2} \sum_{i,j,i\neq j} \left\| \mathbf{U}^{(ij)}\big(s \cdot \mathbf{f}^{(ij)} + \mathbf{g}^{(ij)}\big) \right\|^2 \right\} \quad (18)$$

$$= s \sum_{i,j,i\neq j} \left( \mathbf{f}^{(ij)\mathsf{T}} \mathbf{U}^{(ij)\mathsf{T}} \mathbf{U}^{(ij)} \mathbf{f}^{(ij)} \right) + \sum_{i,j,i\neq j} \left( \mathbf{f}^{(ij)\mathsf{T}} \mathbf{U}^{(ij)\mathsf{T}} \mathbf{U}^{(ij)} \mathbf{g}^{(ij)} \right). \quad (19)$$

Therefore, the scale parameter $s$ is computed by

$$s = - \frac{\sum_{i,j,i\neq j} \left( \mathbf{f}^{(ij)\mathsf{T}} \mathbf{U}^{(ij)\mathsf{T}} \mathbf{U}^{(ij)} \mathbf{g}^{(ij)} \right)}{\sum_{i,j,i\neq j} \left( \mathbf{f}^{(ij)\mathsf{T}} \mathbf{U}^{(ij)\mathsf{T}} \mathbf{U}^{(ij)} \mathbf{f}^{(ij)} \right)}. \quad (20)$$

### 1.3   Projection function $\pi^{(i)}(\cdot)$ for the scale-oriented BA

In the main paper, the projection function $\pi^{(i)}(\cdot)$ of the $l^{\text{th}}$ FIR 3D point $\mathbf{X}_l$ (in the world coordinate system) for the $i^{\text{th}}$ FIR camera is defined by

$$\pi^{(i)}(s, \mathbf{X}_l) = \left[ f_x X_l^{(i)}/Z_l^{(i)} + c_x,\ f_y Y_l^{(i)}/Z_l^{(i)} + c_y \right]^{\mathsf{T}}. \tag{21}$$

Note that $\mathbf{X}_l^{(i)} = [X_l^{(i)}, Y_l^{(i)}, Z_l^{(i)}]^{\mathsf{T}}$ indicates the $l^{\text{th}}$ FIR 3D point in the coordinate system of the $i^{\text{th}}$ FIR camera. In Algorithm (1), the global extrinsic parameter of the $i^{\text{th}}$ FIR camera is computed by

$$\mathbf{T}_{\text{f}}^{(i)} = \begin{bmatrix} \mathbf{R}_{\text{s}} & \mathbf{t}_{\text{s}} \\ \mathbf{0}^{\mathsf{T}} & 1 \end{bmatrix} \begin{bmatrix} \mathbf{R}_{\text{v}}^{(i)} & s \cdot \mathbf{t}_{\text{v}}^{(i)} \\ \mathbf{0}^{\mathsf{T}} & 1 \end{bmatrix} \tag{22}$$

$$= \begin{bmatrix} \mathbf{R}_{\text{s}} \mathbf{R}_{\text{v}}^{(i)} & s \cdot \mathbf{R}_{\text{s}} \mathbf{t}_{\text{v}}^{(i)} + \mathbf{t}_{\text{s}} \\ \mathbf{0}^{\mathsf{T}} & 1 \end{bmatrix}. \tag{23}$$

Using Equation (23), $\mathbf{X}_l^{(i)}$ is calculated as follows:

$$\mathbf{X}_l^{(i)} = \mathbf{R}_{\text{s}} \mathbf{R}_{\text{v}}^{(i)} \mathbf{X}_l + s \cdot \mathbf{R}_{\text{s}} \mathbf{t}_{\text{v}}^{(i)} + \mathbf{t}_{\text{s}} \tag{24}$$

On the contrary in Algorithm (2), $\mathbf{T}_{\text{f}}^{(i)}$ is computed by

$$\mathbf{T}_{\text{f}}^{(i)} = \begin{bmatrix} \mathbf{R}_{\text{s}} & s \cdot \mathbf{t}_{\text{s}} \\ \mathbf{0}^{\mathsf{T}} & 1 \end{bmatrix} \begin{bmatrix} \mathbf{R}_{\text{v}}^{(i)} & \mathbf{t}_{\text{v}}^{(i)} \\ \mathbf{0}^{\mathsf{T}} & 1 \end{bmatrix} \tag{25}$$

$$= \begin{bmatrix} \mathbf{R}_{\text{s}} \mathbf{R}_{\text{v}}^{(i)} & \mathbf{R}_{\text{s}} \mathbf{t}_{\text{v}}^{(i)} + s \cdot \mathbf{t}_{\text{s}} \\ \mathbf{0}^{\mathsf{T}} & 1 \end{bmatrix}. \tag{26}$$

Using Equation (26), $\mathbf{X}_l^{(i)}$ is calculated as follows:

$$\mathbf{X}_l^{(i)} = \mathbf{R}_{\text{s}} \mathbf{R}_{\text{v}}^{(i)} \mathbf{X}_l + \mathbf{R}_{\text{s}} \mathbf{t}_{\text{v}}^{(i)} + s \cdot \mathbf{t}_{\text{s}}. \tag{27}$$

## 2   Synthetic image experiments under noise-free settings

This section explains the result of the synthetic image experiments in the environment where there is no feature point detection error. In the main paper, we introduce the two approaches for the scale estimation, which are called Algorithms (1) and (2). In this section, we confirm that both Algorithms (1) and (2) are capable of estimating the scale parameter $s$ correctly under the noise-free condition with respect to feature point detection.

### 2.1   Experimental settings

In order to confirm both Algorithms (1) and (2) can estimate the correct scale parameter, we first prepare the synthetic dataset whose absolute scale is incorrect, and then estimate the scale parameter of the monocular SfM. Thus, we should reproduce scale ambiguity of monocular SfM in the simulation.

The procedure for the experiment is almost the same as those written in Section 4.1 of the main paper. However, in order to reproduce the scale ambiguity of monocular SfM after the step 3, we scale the positions of the RGB cameras by the scaling ratio $s_{\text{true}}$ as origin reference.

The procedure for the experiment is as follows:

1. Scatter 3D points $\mathbf{X}_i \in \mathbb{R}^3$ ($i = 1, 2, \cdots, n_{\text{p}}$) randomly in a cubic space with a side length of $D$.
2. Arrange $n_{\text{c}}$ RGB–FIR camera systems in the 3D space randomly. More concretely, a constant relative transformation of an RGB–FIR camera system $\mathbf{T}_{\text{s}}$ is given, and the absolute camera poses of the RGB cameras $\mathbf{T}_{\text{v}}^{(k)}$ ($k = 1, 2, \cdots, n_{\text{c}}$) are set randomly, then the absolute camera poses of the FIR cameras $\mathbf{T}_{\text{f}}^{(k)}$ ($k = 1, 2, \cdots, n_{\text{c}}$) are computed by $\mathbf{T}_{\text{f}}^{(k)} = \mathbf{T}_{\text{s}} \mathbf{T}_{\text{v}}^{(k)}$.
3. For all $k = 1, 2, \cdots, n_{\text{c}}$, reproject the 3D points $\mathbf{X}_1, \mathbf{X}_2, \cdots, \mathbf{X}_{n_{\text{p}}}$ to the $k^{\text{th}}$ FIR camera using $\mathbf{T}_{\text{f}}^{(k)}$, then determine the normalized image points $\mathbf{p}_i^{(k)}$ ($i = 1, 2, \cdots, n_{\text{p}}$). The noise is not added to the reprojected points, namely, $\sigma_{\text{n}} = 0$.
4. To reproduce the scale ambiguity of monocular SfM, scale the positions of the RGB cameras by the scaling ratio $s_{\text{true}}$ as origin reference.
5. Estimate the scale parameter $s$ using both Algorithms (1) and (2).

In the experiment, we define $n_{\text{p}} = 1000$, $D = 2000$ and $n_{\text{c}} = 100$. In addition, the relative pose $\mathbf{T}_{\text{s}}$ between the two cameras of the RGB–FIR camera system is set as

$$\mathbf{T}_{\text{s}} = \begin{bmatrix} \mathbf{R}_{\text{s}} & \mathbf{t}_{\text{s}} \\ \mathbf{0}^{\mathsf{T}} & 1 \end{bmatrix} \text{ with } \mathbf{R}_{\text{s}} = \mathbf{I} \text{ and } \mathbf{t}_{\text{s}} = \begin{bmatrix} d & 0 & 0 \end{bmatrix}^{\mathsf{T}}, \qquad (28)$$

where $d = 1.0$ is the distance between the two cameras of the RGB–FIR camera system. We perform the whole procedure (from the step 1 to 5) a hundred times for every $s_{\text{true}}$, then calculate a mean of the estimated scale parameters with 100 trials.

## 2.2  Results of noise-free settings

From the definitions of $\mathbf{T}_{\text{f}}^{(i)}$, it is clear that $s_{\text{true}}$ and $s$ should be the same when using Algorithm (2), whereas $s_{\text{true}}$ and $s^{-1}$ should be the same when using Algorithm (1). For this reason, Fig. 1a shows a plot of $s^{-1}$, which is estimated using Algorithm (1), with respect to $s_{\text{true}}$. On the contrary, Fig. 1b represents a plot of $s$, which is estimated using Algorithm (2), with respect to $s_{\text{true}}$. In both Figs. 1a and 1b, we present the estimated values as a mean with 100 trials of the simulation, which is composed of the preparation of the synthetic dataset and the scale estimation, for each $s_{\text{true}}$.

In Fig. 1a, the reciprocal of the estimated scale $s^{-1}$ is exactly equal to scaling ratio $s_{\text{true}}$ in the whole range of $[10^{-2}, 10^2]$. In addition, the estimated scale $s$ is also equal to the scaling ratio $s_{\text{true}}$ in the whole range of $s_{\text{true}}$ in Fig. 1b. Therefore, we confirmed that both Algorithms (1) and (2) are capable of estimating

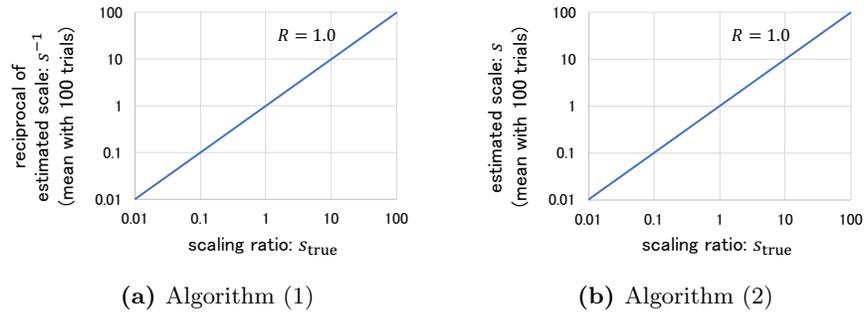

**(a)** Algorithm (1)                                **(b)** Algorithm (2)

**Fig. 1.** Estimated scale $s$ under noise-free settings ($\sigma_\mathrm{n} = 0$) when the scaling ratio $s_\mathrm{true}$ is varied in the range of $[10^{-2}, 10^2]$. The correlation coefficients of both (a) and (b) are $R = 1.0$, that is to say, the scale parameters estimated using both Algorithms (1) and (2) are exactly equal to the correct scales.

the scale parameter $s$ correctly under the noise-free condition with respect to feature point detection.

## 3    Calibration of RGB–FIR camera system

Pre-calibration of an RGB–FIR stereo camera system is needed to perform the proposed scale estimation procedure. Thus, we adopt the calibration method in which a planar pattern such as a chessboard is used [2]. A planar pattern calibration board that can be recognized by FIR sensors is necessary to apply the calibration method above to the RGB–FIR stereo camera system. Being inspired by [1], we produce a grid-aligned visible circle pattern board on which a small heater is embedded at the center of each circle, which can be recognized by FIR cameras as well as by RGB cameras, as shown in Fig. 2a. Intrinsic and extrinsic parameters of an RGB–FIR stereo camera system shown in Fig. 2b are estimated using the method proposed in the work by [2].

## 4    Additional experimental results

This section shows the additional experimental results. In Section 4.1, we show the thermal 3D mappings of the evaluation target scene used in the main paper, as one of the applications of the proposed method. Sections 4.2 and 4.3 present the thermal 3D mappings in the outdoor scene and in the indoor scene, respectively.

### 4.1    Thermal 3D mappings of the target scene for the evaluation

Figs. 3a–3d represent the thermal 3D mappings of the target scene for the evaluation in the main paper, under the four baseline setups: 273[mm], 192[mm], 113[mm] and 26[mm], respectively. Fig. 3e is the RGB mesh model of the scene.

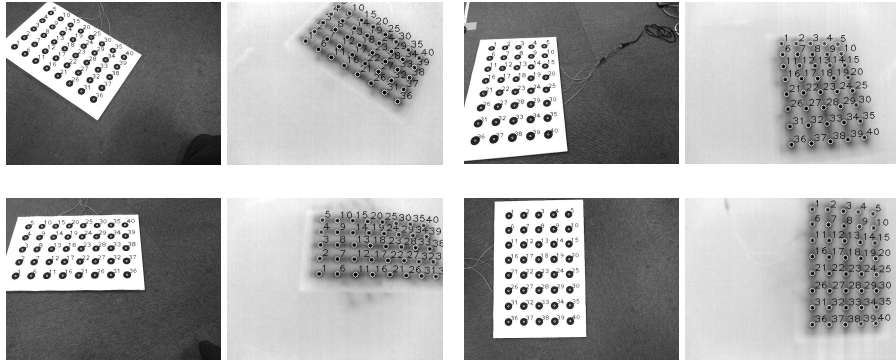

(a) RGB (*left*) and FIR (*right*) images of the calibration board

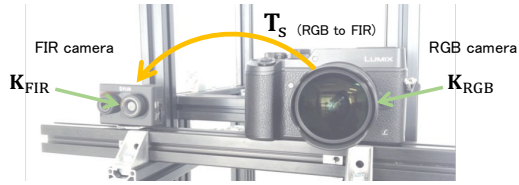

(b) Parameters to calibrate ($\mathbf{T}_s, \mathbf{K}_{\mathrm{RGB}}, \mathbf{K}_{\mathrm{FIR}}$)

**Fig. 2.** (a) RGB and FIR images of the calibration board. The detected circles are numbered with consecutive numbers. (b) Parameters to calibrate. $\mathbf{T}_s$ is the relative transformation between the RGB and FIR cameras, which compose an RGB–FIR camera system. $\mathbf{K}_{\mathrm{RGB}}$ and $\mathbf{K}_{\mathrm{FIR}}$ are the intrinsic parameters of the RGB and FIR cameras, respectively.

The thermal information is reprojected well in Figs. 3a–3d because of the estimated scale parameter with high accuracy. The spacial thermal changes approximately coincide with the boundaries of the 3D shape, such as the edge of the bottle, the monitor, and the electric kettle.

### 4.2 Real image experiments in the outdoor scene

Fig. 4 shows the thermal 3D mapping of the outdoor scene (dataset *facade*). The number of the RGB–FIR image pairs in the dataset *facade* is 1029. The RGB–FIR camera system used in the experiment is a FLIR Duo R (FLIR Systems, Inc), the baseline length of which is 26[mm]. It is observed that the temperature in the shade looks lower than that in the sunny places.

### 4.3 Real image experiments in the indoor scene

Fig. 5 shows the thermal 3D mapping of the indoor scene (dataset *office*). The number of the RGB–FIR image pairs in the dataset *office* is 823. The RGB–FIR camera system used in the experiment is composed of a LUMIX DMC–G8

(Panasonic Corp.) as the RGB camera and a FLIR Duo R (FLIR Systems, Inc) as the FIR camera. The baseline length of the RGB–FIR camera system is 242[mm].

## References

1. Weinmann, M., Leitloff, J., Hoegner, L., Jutzi, B., Stilla, U., Hinz, S.: Thermal 3d mapping for object detection in dynamic scenes. ISPRS Annals of the Photogrammetry, Remote Sensing and Spatial Information Sciences **2**(1),  53 (2014)
2. Zhang, Z.: A flexible new technique for camera calibration. IEEE Transactions on Pattern Analysis and Machine Intelligence (TPAMI) **22**, 1330–1334 (2000)

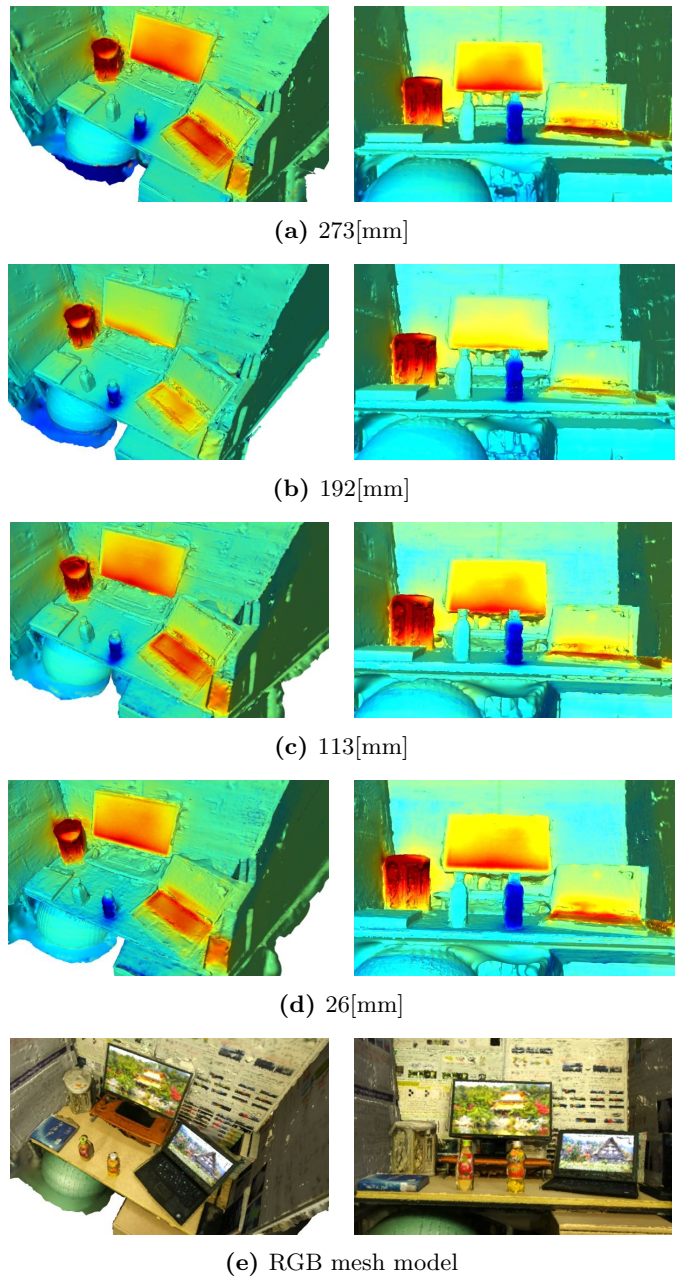

**(a)** 273[mm]

**(b)** 192[mm]

**(c)** 113[mm]

**(d)** 26[mm]

**(e)** RGB mesh model

**Fig. 3. (a)**, **(b)**, **(c)**, **(d)** Thermal 3D mappings of the target scene for the evaluation in the main paper, under the four baseline setups: 273[mm], 192[mm], 113[mm] and 26[mm], respectively. It is found that the thermal information is reprojected well thanks to the accurate scale parameter estimated by the proposed method. **(e)** RGB mesh model of the scene.

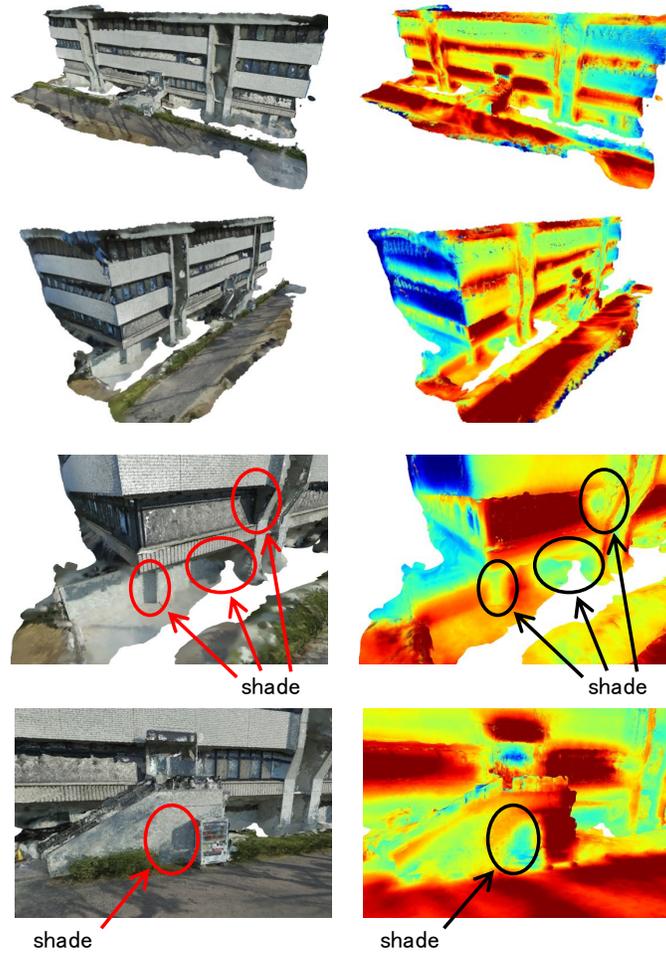

**Fig. 4.** Thermal 3D mapping of the outdoor scene (dataset *facade*). The number of the RGB–FIR image pairs in the dataset *facade* is 1029. The RGB–FIR camera system used in the experiment is a FLIR Duo R (FLIR Systems, Inc), the baseline length of which is 26[mm]. It is found that the temperature in the shade looks lower than that in the sunny places.

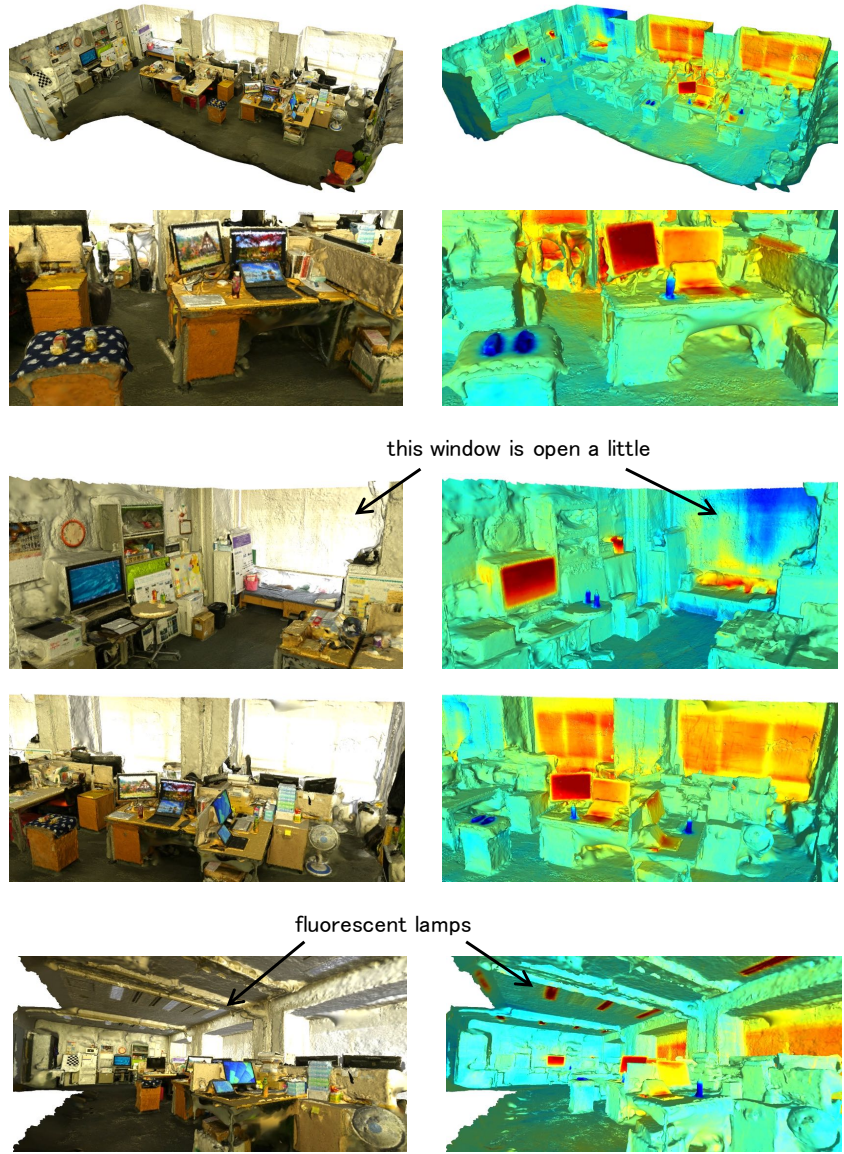

**Fig. 5.** Thermal 3D mapping of the indoor scene (dataset *office*). The number of the RGB–FIR image pairs in the dataset *office* is 823. The RGB camera in the RGB–FIR camera system used in the experiment is a LUMIX DMC–G8 (Panasonic Corp). The FIR camera in the camera system is a FLIR Duo R (FLIR Systems, Inc). The baseline length of the RGB–FIR camera system is 242[mm].



\end{document}